\documentclass[times, review, 10pt]{elsarticle}

\usepackage{lineno,hyperref}
\modulolinenumbers[5]
\usepackage{times}
\usepackage{epsfig}
\usepackage{graphicx}
\usepackage{amsmath}
\usepackage{amssymb}  
\usepackage{stfloats}
\usepackage{bm}
\usepackage{multirow}
\usepackage{latexsym}
\usepackage{threeparttable}
\usepackage{xcolor}
\usepackage{footmisc}
\usepackage{makecell}
\usepackage{bbding}
\usepackage{framed}
\usepackage{subfigure}
\usepackage{booktabs}
\usepackage{algorithm}
\usepackage{algorithmic}
\usepackage{tablefootnote}
\usepackage[bottom=4.3cm,top=4.3cm,left=4.8cm,right=4.8cm]{geometry}
\usepackage[algo2e,ruled,vlined]{algorithm2e}

\journal{Pattern Recognition}

\let\oldequation\equation
\let\oldendequation\endequation

\renewenvironment{equation}
{\linenomathNonumbers\oldequation}
{\oldendequation\endlinenomath}

\begin{document}

\begin{frontmatter}

%
%
%
%
%
%

\title{Learning Contrastive Feature Representations for Facial Action Unit Detection}

\author[mymainaddress1,mymainaddress2]{Ziqiao Shang}
\ead{ziqiaosh@gmail.com}

\author[mymainaddress3]{Bin Liu\corref{cor1}}
\ead{binliu@swjtu.edu.cn}

\author[mymainaddress3]{Fengmao Lv}
\ead{fengmaolv@126.com}

\author[mymainaddress3]{Fei Teng}
\ead{fteng@swjtu.edu.cn}

\author[mymainaddress3]{Tianrui Li}
\ead{trli@swjtu.edu.cn}

\author[mymainaddress1,mymainaddress2]{Lan-Zhe Guo}
\ead{guolz@lamda.nju.edu.cn}

\cortext[cor1]{Corresponding author.}
\address[mymainaddress1]{National Key Laboratory for Novel Software Technology, Nanjing University, Nanjing, China.}
\address[mymainaddress2]{School of Intelligence Science and Technology, Nanjing University, Suzhou, China.}
\address[mymainaddress3]{School of Computing and Artificial Intelligence, Southwest Jiaotong University, Chengdu, China.}

\begin{abstract}
	For the Facial Action Unit (AU) detection task, accurately capturing the subtle facial differences between distinct AUs is essential for reliable detection. Additionally, AU detection faces challenges from class imbalance and the presence of noisy or false labels, which undermine detection accuracy. In this paper, we introduce a novel contrastive learning framework aimed for AU detection that incorporates both self-supervised and supervised signals, thereby enhancing the learning of discriminative features for accurate AU detection. To tackle the class imbalance issue, we employ a negative sample re-weighting strategy that adjusts the step size of updating parameters for minority and majority class samples. Moreover, to address the challenges posed by noisy and false AU labels, we employ a sampling technique that encompasses three distinct types of positive sample pairs. This enables us to inject self-supervised signals into the supervised signal, effectively mitigating the adverse effects of noisy labels. Our experimental assessments, conducted on five widely-utilized benchmark datasets (BP4D, DISFA, BP4D+, GFT and Aff-Wild2), underscore the superior performance of our approach compared to state-of-the-art methods of AU detection. Our code is available at \url{https://github.com/Ziqiao-Shang/AUNCE}.
\end{abstract}





\end{frontmatter}


\section{Introduction}
Facial Action Coding System (FACS)~\cite{rosenberg:et.al:2020} is a widely employed approach for facial expression coding, which introduced a set of 44 facial action units (AUs) that establish a link between facial muscle movements and facial expressions~\cite{fathallah2017facial}. AU detection is a multi-label binary classification problem. Each AU represents a kind of label, and activated and inactivated AUs are respectively labeled 1 and 0. Nowadays, AU detection~\cite{shao2025facial} has emerged as a computer vision technique to automatically identify AUs in videos or images. This advancement holds promise for diverse applications, including human-computer interaction, emotion analysis, car driving monitoring et al.

In recent years, supervised learning has been the dominant paradigm for developing AU detection methods~\cite{shao2026constrained}. A common practice in these methods is to use binary cross-entropy (BCE) loss as the optimization objective. However, BCE loss inherently prioritizes strict label alignment and fine-grained pixel-level feature encoding during training. This tendency can lead to two key limitations: first, it may cause the model to overfit to training data (as it overemphasizes minor label or pixel variations that are irrelevant to AU detection); second, it often increases overall model complexity by forcing the network to learn overly detailed features—even though subtle differences in AU-related patterns are usually sufficient for accurate detection. Together, these issues restrict the model’s ability to generalize to unseen data, hindering its performance in real-world scenarios.

\begin{figure}[t!]
	\centering
	\setlength{\abovecaptionskip}{0.05cm}
	\includegraphics[width=0.99\linewidth]{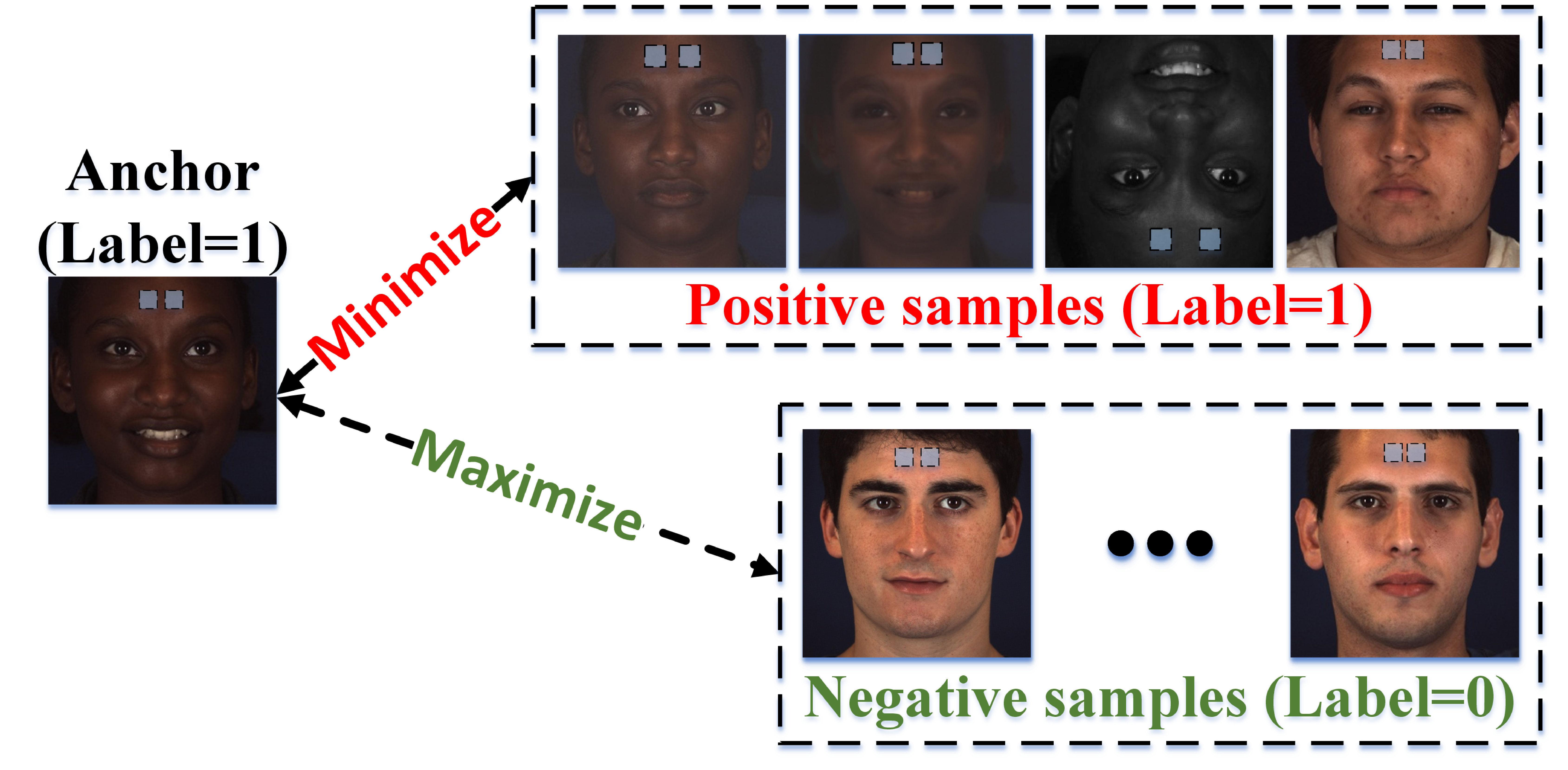}
	\caption{Illustration of discriminative contrastive learning frameworks for representation learning. The goal is to maximize the similarity between positive samples (semantically similar) and minimize the similarity between negative samples (semantically dissimilar).}
	\label{fig:1}
\end{figure}

As a result, representation learning approaches based on discriminative contrastive learning have emerged and shown significant potential~\cite{assran2023self} for classification tasks. As shown in Fig.~\ref{fig:1}, this approach aims to maximize the consistency of positive samples that are semantically similar while minimizing the consistency of negative samples that are semantically dissimilar. By optimizing the contrastive loss, the mutual information between variables is either explicitly or implicitly optimized~\cite{henaff2020data}. However, the direct integration of contrastive learning into AU detection encounters two principal challenges. The first challenge is the class imbalance issue of each AU type~\cite{zhou2020bbn}. Most AU classes manifest a substantial imbalance between majority and minority class samples, with inactivated AUs vastly outnumbering their activated counterparts. This imbalance issue influences the classification decision threshold, potentially resulting in the direct classification of the majority of AUs as the negative class. The second challenge pertains to the noisy and false label problem~\cite{li2024contrastive}. Due to the subtle feature differences when AUs activate leading to the inherent difficulty in expertly labeling AUs, existing AU datasets may exhibit a few noisy and false labels, significantly undermining the overall fitting ability of models.

To address these challenges, we propose a discriminative contrastive learning framework inspired by self-supervised learning techniques for learning highly semantic image representations~\cite{assran2023self}. This framework encodes differential information between AUs from each sample pair, offering an alternative to the traditional pixel-level feature learning methods commonly employed in AU detection. The resulting loss, termed AUNCE, is conceptually similar to the widely used InfoNCE loss. Moreover, to tackle the class imbalance issue in AU detection, we adopt a negative sample re-weighting strategy inspired by \cite{shang2024facial}. This strategy addresses class imbalance by assigning varying levels of importance to individual samples during training, achieved by adjusting the weights of their loss contributions. Additionally, to mitigate the effects of noisy and false AU labels, inspired by the sample-mixing concept proposed in \cite{zhang2017mixup}, we introduce a positive sample sampling strategy. This strategy enhances supervision signals by incorporating self-supervised signals, achieved by defining sampling ratios for different types of positive sample pairs. This ensures that more robust positive samples are derived through data augmentation combined with self-supervised signal. The primary contributions of our work are summarized as follows:

\begin{itemize}
	\item We propose a discriminative contrastive learning framework that encodes differential information between AUs from each sample pair, and the resulting loss is named AUNCE.
	\item We adopt a negative sample re-weighting strategy to tackle the class imbalance issue of each AU type.
	\item We introduce a positive sample sampling strategy to mitigate the effects of noisy and false AU labels.
	\item We validate the effectiveness of AUNCE by comparing it to state-of-the-art approaches for AU detection on five widely-used datasets, namely BP4D, DISFA, BP4D+, GFT and Aff-Wild2.
\end{itemize}

\section{Related Work}
We provide a comprehensive overview of two areas closely connected to our research: discriminative contrastive learning and contrastive learning for AU detection.
\subsection{Discriminative contrastive learning} 
Contrastive learning~\cite{henaff2020data} originated in self-supervised learning, where tasks are assigned to agents by applying various augmentations to the same image and evaluating feature similarity. The goal is to maximize similarity between features of the same image while minimizing it between different images. Over time, contrastive learning has expanded to supervised and semi-supervised settings, leveraging labeled data to refine feature representations.

In recent years, discriminative contrastive learning, a subset of deep self-supervised learning methods~\cite{chen2020simple,he2020momentum}, has witnessed substantial progress. This approach follows a learn-to-compare paradigm, distinguishing between similar (positive) and dissimilar (negative) data points in a feature space. For instance, Oord \textit{et al.}\cite{henaff2020data} introduced the InfoNCE loss to optimize feature extraction by maximizing consistency in sequence data, establishing a foundation for contrastive methods. Chen \textit{et al.}\cite{chen2020simple} proposed SimCLR, which uses a base encoder and projection head to maximize agreement through InfoNCE. Similarly, MoCo~\cite{he2020momentum} leverages a momentum network to maintain a queue of negative samples for efficient learning. While variations in representation encoders and similarity measures exist, the core principle remains: training a representation encoder $f$ by contrasting similar pairs ($x, x^{+}$) against dissimilar pairs ($x, x^{-}$) to optimize the contrastive loss.

In addition to enabling the automatic extraction of fine-grained feature representations via the comparison of positive and negative pairs, deep contrastive learning has evolved to incorporate diverse contrastive objectives tailored for different supervision levels. In this context, Khosla \textit{et al.}~\cite{khosla2020supervised} extended the contrastive learning concept to supervised settings with a supervised contrastive learning loss, aiming to enhance feature expression using labeled data. Additionally, Robinson \textit{et al.}~\cite{robinson2021contrastive} introduced a label-free strategy to sample hard negatives in contrastive learning, enabling tighter class clustering and improved downstream performance with minimal implementation overhead.

In discriminative contrastive learning, the construction of positive sample pairs has evolved. For labeled or partially labeled data, positive pairs can be formed using images from the same category, increasing pair diversity and enhancing feature representation. Contrastive learning also supports supervised learning by enabling pretraining that boosts downstream performance. However, applying an InfoNCE-based contrastive learning framework to AU detection still faces many significant challenges, including class imbalance, noisy annotations, and the difficulty of distinguishing hard negatives.

\subsection{Contrastive learning for AU detection}
Contrastive learning was firstly designed as a pretext task for unsupervised or self-supervised learning and was introduced to the domain of AU detection with limited annotations~\cite{niu2019multi,zhang2023weakly}. For example, Niu \textit{et al.}~\cite{niu2019multi} introduced a semi-supervised co-training approach that utilized contrastive loss to enforce conditional independence between facial representations generated by two Deep Neural Networks (DNNs). Following this, Zhang \textit{et al.}~\cite{zhang2023weakly} introduced a weakly-supervised contrastive learning approach that effectively utilizes coarse-grained activity information.

Subsequent research endeavors extended contrastive learning to supervised settings, aiming at enhancing feature representation for AU detection through the utilization of labeled data~\cite{chang2022knowledge,suresh2022using}. Chang \textit{et al.}~\cite{chang2022knowledge} present a knowledge-driven representation learning framework aimed for AU detection, incorporating a contrastive learning module that leverages dissimilarities among facial regions to effectively train the feature encoder. In a similar vein, Suresh \textit{et al.}~\cite{suresh2022using} proposed a feature-based positive matching contrastive loss, which facilitates learning distances between positives based on the similarity of corresponding AU embeddings.

However, previous methods primarily employed a combination of weighted cross-entropy loss and triplet loss to constrain the training process, encoding pixel-level information of the entire face associated with AUs. This results in significant redundant information that may dilute the discriminative signals necessary for accurate AU detectionTo address this, Sun \textit{et al.}~\cite{sun2021emotion} proposed EmoCo, an unsupervised contrastive learning framework that leverages emotion-based image pairs. Although EmoCo improves AU detection performance, it suffers from label inconsistencies within emotion categories, leading to false positives and negatives, and its dependence on emotion annotations restricts its applicability to datasets lacking such labels. Similarly, Li \textit{et al.}~\cite{li2024contrastive} introduced a self-supervised contrastive approach based on the MoCo framework~\cite{he2020momentum} for dynamic AU detection in videos. However, its reliance on temporal cues limits its generalization to static images. To further address cross-domain generalization, Li \textit{et al.}~\cite{li2025decoupled}, which combines image- and feature-level contrastive learning to disentangle AU-relevant and domain-specific representations. Despite its effectiveness, its performance may decline under large domain shifts or limited AU data diversity. 

In summary, while these approaches have significantly advanced the field of AU detection, each is confronted with unique challenges, including the encoding of redundant pixel-level information, reliance on emotion annotations, and limitations in handling static image datasets. Furthermore, there is currently no effective solution for applying discriminative contrastive learning methods to static image datasets that contain only AU annotations. This gap underscores the need for more adaptable and efficient strategies in AU detection.

\begin{figure*}[t!]
	\begin{center}
		\setlength{\abovecaptionskip}{0.05cm}
		\includegraphics[width=0.99\linewidth]{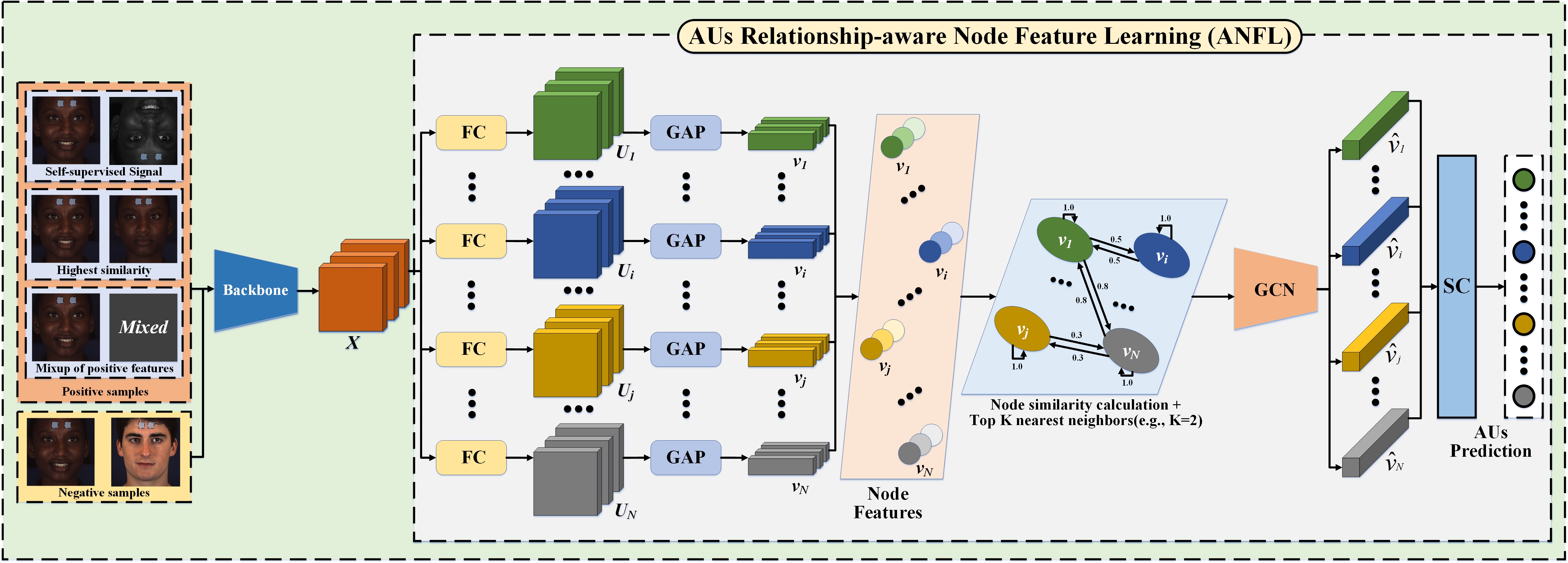}
		\caption{Overview of the training pipeline of our discriminative contrastive learning framework. The training process consists of two stages: pretraining and linear evaluation. The pretraining stage precedes the linear evaluation stage. In the first stage, the feature encoder is pretrained by the AUNCE loss. In the second stage, we adopt a linear evaluation protocol, consistent with the practice established in contrastive learning frameworks such as SimCLR. Specifically, a single linear fully connected layer is trained atop the frozen encoder to assess the quality of the learned feature representations.}
		\label{fig:2}
	\end{center}
\end{figure*}

\section{Method} \label{method}
\subsection{Objectives} \label{3-1}
In this study, we confront three main challenges: 1. How to design a novel framework that guides the encoder to focus on the differential information between AUs from each sample pair \cite{sun2021emotion}, going beyond traditional methods reliant on pixel-level facial features, 2. How to tackle the class imbalance issue across different AU types \cite{zhou2020bbn} to improve the overall accuracy of our framework. 3. How to handle noisy and false labels \cite{li2024contrastive} to increase the model's robustness against misannotations.

To address these challenges, Section~\ref{method} provides detailed solutions. In Section~\ref{3-2}, we propose a discriminative contrastive learning framework that directs the encoder to capture differential features between AUs from each sample pair, thus moving away from conventional pixel-level feature learning. To combat the class imbalance issue, Section~\ref{sec:neg} introduces a negative sample re-weighting strategy, drawing inspiration from, which prioritizes underrepresented AUs and facilitates the detection of challenging samples. Besides, Section~\ref{sec:pos} presents a positive sample sampling strategy that incorporates augmented supervision signals, which helps the model effectively learn from noisy and false data. Finally, Section~\ref{sec:encoder} will introduce the encoder structure we used in our framework.

\subsection{Discriminative contrastive learning framework} \label{3-2}
For the task of AU detection, which falls under the category of multi-label binary classification, the prevalent approach for training models has been the utilization of cross-entropy loss. This loss measures the disparity between predicted labels and ground truth labels. However, there exists variability in the occurrence rates of different AUs, with some AUs exhibiting low occurrence rates while others display high occurrence rates. Consequently, in contrast to the conventional cross-entropy loss, we employ a weighted cross-entropy loss~\cite{shao2021jaa} denoted as ${L}_{wce}$
\begin{equation} \label{eq:wce} 
	\begin{aligned}
		&{L}_{wce}=-\frac{1}{n_{au}}\sum_{i=1}^{n_{au}}w_{i}[y_{i}log\hat{y}_{i}+(1-{y}_{i})log(1-\hat{y}_{i})],
	\end{aligned}
\end{equation}
where $y_{i}$ denotes the ground truth value of occurrence (\textit{i.e.}, 0 or 1) of the $i$-th AU, and $\hat{y}_{i}$ is the corresponding predicted occurrence probability. This loss effectively guides the training process, encouraging the model to allocate higher probabilities to the correct class labels and lower probabilities to incorrect ones. $w_{i}$=$\frac{(1/r_{i})n_{au}}{\sum_{i=1}^{n_{au}}(1/r_{i})}$ is the weight of the $i$-th AU determined by its occurrence frequency for balanced training. It permits reducing the impact of loss values associated with those AUs that are more commonly activated in the training set. $r_{i}$ represents the occurrence frequency of the $i$-th AU, and $n_{au}$ is the total number of AUs. 

Though cross-entropy loss is widely used and effective, it has limitations in tasks with class imbalance, complex data distributions, or fine-grained distinctions~\cite{wei2021fine}. It can lead to overfitting and poor generalization, particularly without regularization or techniques like label smoothing. In AU detection, where subtle inter-class differences are critical, cross-entropy loss may restrict the model's ability to generalize to unseen data. Additionally, the requirement for predicted outputs $\hat{y}$ to satisfy specific conditions for each AU imposes a significant burden on the encoder's learning capacity. 

To address this challenge, we introduce a discriminative contrastive learning framework for AU detection. As shown  in Fig.\ref{fig:2}, the training process comprises two stages: pretraining and linear evaluation. The pretraining stage, preceding linear evaluation, integrates a feature encoder with two key components—a positive sample mining module and a negative sample re-weighting module. The former combines self-supervised and supervised signals to mitigate the impact of noisy labels caused by subtle AU variations, while the latter alleviates class imbalance and enhances hard negative sample mining. Further details are provided in Sections~\ref{sec:neg} and~\ref{sec:pos}.

The input to the encoder consists of an anchor point $x$, a positive sample $x^+$ that belongs to the same class as $x$, and negative samples $x^-$ from different classes within the same batch. These samples are passed through a shared encoder to extract feature representations. Finally, the encoder parameters are optimized using a novel contrastive loss, referred to as AUNCE, which is defined as follows:

\begin{equation}\label{eq:aunce1} 
	\begin{aligned}
		&\mathcal{L}_{aunce} = \mathbb{E}_{\substack{{x}\sim{p}_{d} \\ {x}^{+}\sim{p}^{+} \\ {x}^{-}\sim{p}^{-}}}\{-\frac{1}{n_{au}}\sum_{i=1}^{n_{au}}w_{i}\\
		&[-\log \frac{\text{sim}(f(x),f(x^+))}{\text{sim}(f(x),f(x^+)) + \sum_{j=1}^{N} \omega_j \cdot \text{sim}(f(x),f(x^-_j))}]\},
	\end{aligned}
\end{equation}

where $f(\cdot)$ serves as an encoder, mapping the features to a $d$ dimensional vector, represents the feature representation of a specific AU. $\text{sim}()$ represents the inner product similarity between two tensors. Specifically, given two tensors $\mathbf{u}$ and $\mathbf{v}$, $\text{sim}(\mathbf{u}, \mathbf{v})$ is computed as their dot product, defined by $\mathbf{u}^\top \mathbf{v}$. $\text{sim}(f(x),f(x^+))$ measures the similarity between $x$ and its positive example $x^+$, while $\text{sim}(f(x),f(x^-_j))$ measures the similarity between the sample and its negative example. The term ${p}_{d}$ represents the data distribution, ${p}^{+}$ and ${p}^{-}$ respectively denote the class conditional distribution of positive and negative samples of $x$. Inspired by Eq~\eqref{eq:wce}, we introduce $w_{i}$ in Eq~\eqref{eq:aunce1} to mitigate the influence of loss values related to commonly activated AUs in the training set. Additionally, we incorporate $\omega_j$ to re-weight the significance of each negative sample.

The choice of ${L}_{aunce}$ over ${L}_{wce}$ in the pretraining phase is motivated by two key considerations. First, ${L}_{wce}$ relies on explicit label supervision and requires encoding pixel-level information to align with class labels, which imposes a heavier burden on the encoder. In contrast, ${L}_{aunce}$ operates on feature differences between sample pairs without requiring explicit labels, encouraging the encoder to capture discriminative cues between AUs more efficiently. Second, ${L}_{wce}$ constrains the absolute predicted probabilities, which may lead to overfitting and demand greater model capacity. Conversely, contrastive loss optimizes relative similarity between positive and negative pairs without enforcing exact values. This is because the minimal value of $\mathcal{L}_{aunce}$ is 0, which is achieved when $\text{sim}((f(x),f(x^+))-\text{sim}(f(x),f(x^-_j)) \rightarrow +\infty$  for all negative sample index $j \in {1,2,\cdots,N}$.

In the linear evaluation phase, we follow a standard protocol commonly adopted in prior work~\cite{henaff2020data, chen2020simple}, where a linear classifier is trained atop the frozen encoder using weighted cross-entropy loss (WCE). Notably, this classifier is a single fully connected layer, which serves solely to evaluate representation quality, without influencing feature learning. Thus, the model’s performance primarily reflects the effectiveness of AUNCE in guiding representation learning.


\subsection{Negative sample re-weighting strategy} \label{sec:neg}
To address the second challenge posed by class imbalance, we implement a negative sample re-weighting strategy specifically targeting the anchor data points. This also contributes to hard negative sample mining. Inspired by HCL~\cite{robinson2021contrastive}, we apply the following weighting scheme to each negative sample $x_j^-$:
\begin{eqnarray}\label{eq:wei}
	\omega_j = \text{sim}(f(x),f(x^-_j))^{\beta},
\end{eqnarray}
where $\beta \in (0, +\infty)$ can be viewed as the importance weight if we assume that the negative samples follow a von Mises-Fisher (vMF) distribution. The resulting AUNCE loss is:

\begin{small}
	\begin{equation} \label{eq:aunce}
		\begin{aligned}
			&\mathcal{L}_{aunce} = \mathbb{E}_{\substack{{x}\sim{p}_{d} \\ {x}^{+}\sim{p}^{+} \\ {x}^{-}\sim{p}^{-}}}\{-\frac{1}{n_{au}}\sum_{i=1}^{n_{au}}w_{i}\\
			&[-\log \frac{\text{sim}(f(x),f(x^+))}{\text{sim}(f(x),f(x^+)) + \sum_{j=1}^{N} \text{sim}(f(x),f(x^-_j))^{\beta+1}}]\}.
		\end{aligned}
	\end{equation}
\end{small}

Next, we analyze how this weighting scheme addresses the class imbalance issue and facilitates the mining of hard negative samples. Without loss of generality, we consider the similarity calculation as $\text{sim}(f(x),f(x^+)) = \exp(f(x)^Tf(x^+)/\tau)$, where $\tau$ is the temperature coefficient. The inclusion of importance weights transforms the gradient of  $\mathcal{L}$  w.r.t  similarity scores of positive pairs into the following form:
\begin{eqnarray}\label{eq:grad1}
	\frac{\partial\mathcal{L}}{\partial \text{sim}(f(x),f(x^+))}
	=	-\frac{1}{\tau} \sum_{j=1}^N P_j,
\end{eqnarray}
where $P_j=\frac{\exp(f(x)^T f(x^-_j)/\tau )^\beta}{\exp(f(x)^Tf(x^+)/\tau) + \sum_{j=1}^{N} \exp(f(x)^Tf(x^-_j)/\tau)^\beta}$, and  the gradient of similarity scores of negative pairs $\text{sim}(f(x),f(x^-))$ w.r.t $\mathcal{L}$ is
\begin{eqnarray}\label{eq:grad2}
	\frac{\partial\mathcal{L}}{\partial \text{sim}(f(x),f(x^-))}
	=\frac{1}{\tau}  P_j.
\end{eqnarray}

Therefore, the ratio between the gradient magnitude of negative examples and positive examples is given by:
\begin{small}
	\begin{eqnarray}\label{eq:bol}
		|\frac{\partial\mathcal{L} / \partial \text{sim}(f(x),f(x^-))}{\partial\mathcal{L}/\partial \text{sim}(f(x),f(x^+))} |= \frac{\exp(f(x)^Tf(x^-_j)/\tau)^\beta }{  \sum_{j=1}^{N} \exp(f(x)^Tf(x^-_j)/\tau)^\beta} .
	\end{eqnarray}
\end{small}

Eq.~\ref{eq:bol} follows the Boltzmann distribution. As the value of $\beta$ increases, the entropy of the distribution decreases significantly, leading to a sharper distribution in the region of high similarity. This has two consequences: 1. It imposes larger penalties on hard negative samples that are in close proximity to the anchor data point, enabling the model to learn more precise decision boundaries from subtle feature differences when AUs activate. 2. It controls the relative contribution of negative samples to the parameter updates, providing a means to address class imbalance issues.

The primary distinction between our weighting strategy and that of HCL~\cite{robinson2021contrastive} lies in the deliberate omission of mean normalization in Eq.~\eqref{eq:wei}, which enables more precise and adaptive mitigation of class imbalance in AU detection. Specifically, we set $\beta \geq 1$ when the anchor sample belongs to the majority class, amplifying the gradient contributions of minority class samples. Conversely, when the anchor is from the minority class, $\beta \leq 1$ is used to suppress the influence of majority class samples. By tuning the weighting factor $\beta$, the model dynamically adjusts the relative importance of positive and negative samples, thereby enhancing its focus on underrepresented classes. In contrast, the inclusion of mean normalization would entangle the effect of $\beta$ with the normalization term, compromising interpretability and hindering precise gradient modulation. This design choice is therefore critical to achieving effective imbalance-aware learning.


\subsection{Positive sample sampling strategy}\label{sec:pos}
\begin{figure}[t!]
	\centering
	\setlength{\abovecaptionskip}{0.05cm}
	\includegraphics[width=0.99\linewidth]{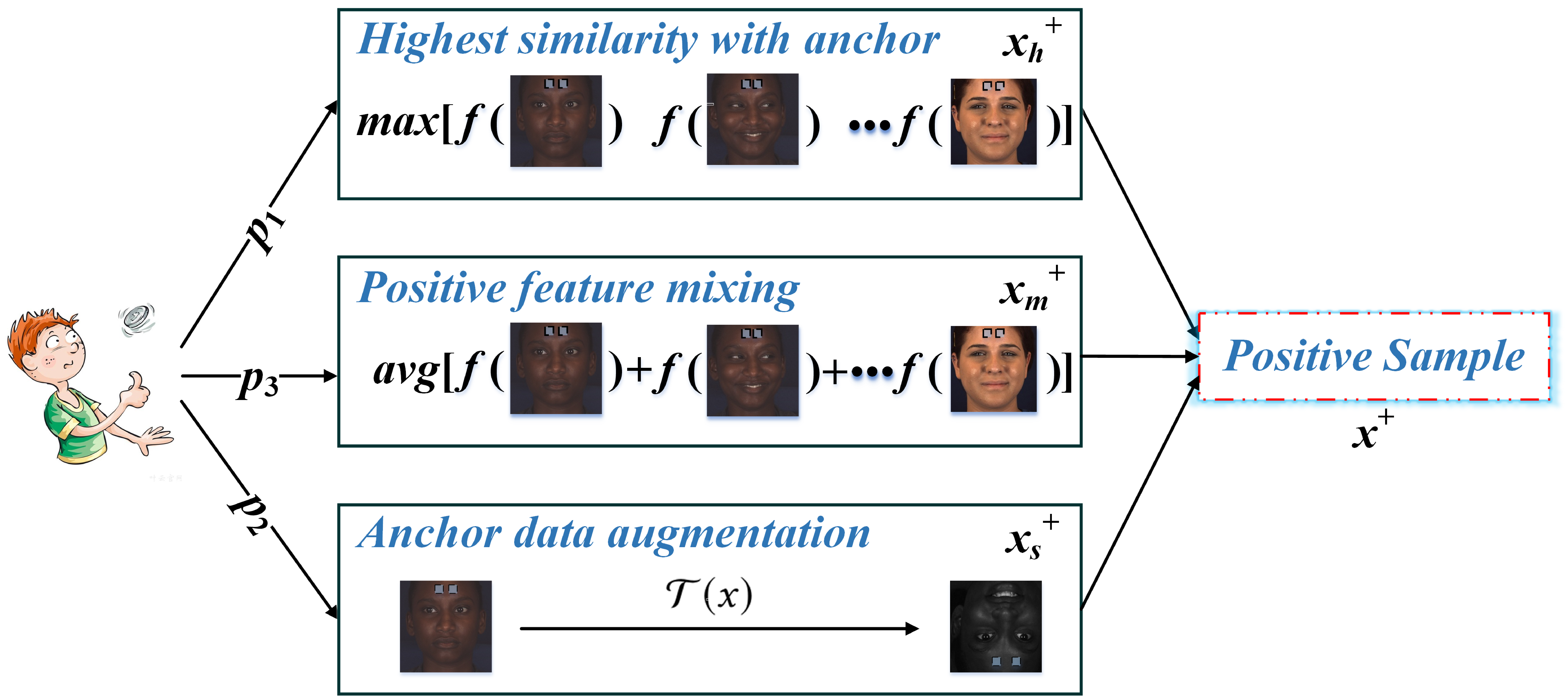}
	\caption{Illustration of our proposed positive sample sampling strategy. It demonstrates the sampling strategy used to mitigate the impact of noisy labels by integrating self-supervised and supervised signals. Positive samples are categorized into three types and sampled with varying probabilities, enhancing the model's robustness to variations and errors in the labeled dataset.}
	\label{fig:3}
\end{figure}

To address the third challenge of AU labels tainted by noise and errors, we adopt a hybrid strategy that combines label noise learning with self-supervised techniques. As illustrated in Fig.~\ref{fig:3}, this method refines the supervision of positive samples by jointly leveraging self-supervised signal and supervised signals, thereby improving model robustness against annotation errors and inconsistencies.

\subsubsection{Supervised Signal} Label smoothing is a widely used technique to mitigate label noise by introducing uncertainty through softened target distributions. However, its direct application in contrastive learning is non-trivial due to the absence of explicit labels in loss computation. To address this, we leverage the memory effect of neural networks—where models tend to learn clean samples before overfitting noisy ones. This phenomenon, observed across architectures, implies that positive samples with higher similarity to the anchor are more likely to be reliably learned.

Inspired by this observation, for a given sample $x$ and M positive samples ${x}_{i=1}^M$ with noisy labels belonging to the same class, we calculate the similarity scores between the representation of sample $x$ and the representations of all M positive samples. We then select the positive feature representation with the highest similarity to sample $x$:
\begin{eqnarray}\label{eq:aug1}
	f(x_{h}^+) = f(x_i), \text{where}~ i = \arg\max_{i} f(x)^Tf(x_i).
\end{eqnarray}

By selecting the positive samples with the highest similarity, we prioritize clean and reliable positive samples while minimizing the influence of noisy positive samples during the training process.
\subsubsection{Self-supervised Signal}
In addition to label noisy learning, we incorporate self-supervised learning techniques into the training process. By designing these auxiliary tasks carefully, we can encourage the model to learn more robust and invariant representations, which can help mitigate the effects of label noise and improve generalization performance. Specifically, our second method to obtain positive feature involves performing semantically invariant image augmentation on the anchor point $x$:
\begin{eqnarray}\label{eq:aug2}
	f(x_{s}^+) = f(\mathcal{T} (x)),
\end{eqnarray}
where $\mathcal{T}(\cdot)$ is a stochastic data augmentation module that applies random transformations to a given data example $x$, generating two correlated views referred to as $x_{s1}^+$ and $x_{s2}^+$, which we consider as a positive pair. Follow SimCLR~\cite{chen2020simple}, we employ a sequential application of three simple augmentations: random cropping and flipping followed by resizing back to original size, random color distortions, and random Gaussian blur. 

Our third method to obtain positive examples involving a mixture of positive feature representations, which can be seen as a mixup of positive samples to obtain the most representative positive instances:
\begin{eqnarray}\label{eq:aug3}
	f(x_{m}^+) = \frac{1}{M} \sum_{i=1}^M f(x_i).
\end{eqnarray}

Unlike data augmentation-based methods such as SimCLR~\cite{chen2020simple}, which rely on instance-level contrast using two augmented views of the same sample, our approach constructs positive examples as class prototypes—defined by the mean representation of multiple same-class samples. This design captures more stable and representative class-level semantics, enabling more robust feature comparisons. The resulting positive feature $f(x_{m}^+)$, as defined in Eq.~\eqref{eq:aug3}, serves as a virtual class centroid synthesized from M positive samples.

The three types of positive examples are constructed by combining label noise learning (for supervised signals) and data augmentation (for self-supervised signals), and are randomly sampled with probabilities $p_1$, $p_2$,and $p_3$ ($p_1+p_2+p_3=1$). This strategy enhances the quality of supervision during training and reduces the influence of noisy labels, thereby improving model robustness and generalization.

By addressing the above three challenges, our approach improves the encoding capability of the shared encode, ultimately leading to improved generalization performance in AU detection. The pseudocode for the algorithm is provided in Algorithm.~\ref{alg:aunce}.

\begin{algorithm}[hbtp]
	\caption{The proposed AUNCE algorithm} \label{alg:aunce}
	\KwIn{Mini batch data $\mathcal{R}$, feature representation function $f(\cdot)$, $p_1$, $p_2$, and $p_3$, $\beta$ .}
	\KwOut{Parameters $\Theta$.}
	\For{\text{sample} $x \in \mathcal{R}$}{
		~~Obtain positive sample set $\{x_i^+\}_{i=1}^M \in \mathcal{R}$;\\
		~~Obtain negative sample set $\{x_j^-\}_{j=1}^N \in \mathcal{R}$;\\
		~~p = random.uniform(0,1); \\
		\If{$p\leq p_1$}{Obtain $f(x_{h}^+)$ via Eq~\eqref{eq:aug1}} 
		~~\eIf{$p_1< p\leq p_1 + p_2$}{Obtain $f(x_{s}^+)$ via Eq~\eqref{eq:aug2}} 
		{Obtain $f(x_{m}^+)$ via Eq~\eqref{eq:aug3}} 
		~~Calculate AUNCE loss via Eq~\eqref{eq:aunce};\\
		~~Update $\Theta$.
	}
	\KwResult{Parameters $\Theta$.}
\end{algorithm}

\subsection{Encoder for AU detection} \label{sec:encoder}
To preserve the fairness of the comparison with conventional pixel-level feature learning methods, our encoder is consistent with the network structure of the first training stage in MEGraph~\cite{luo2022learning}, consisting of Swin-transformer backbone and AU Relationship-aware Node Feature Learning module. The input to the encoder consists of an contrastive larning image pair consisting of an anchor and a random sample, and the encoder is optimized by AUNCE only.

\section{Evaluation}
\subsection{Experiment Setting}
\subsubsection{Datasets}
In our experiments, we extensively utilize four well-established datasets in constrained scenarios for AU detection, namely BP4D~\cite{zhang:et.al:2014:IVC}, DISFA~\cite{mavadati:et.al:2013:TAC}, BP4D+~\cite{zhang2016multimodal},  GFT~\cite{girard2017sayette}, and one well-established dataset in unconstrained scenarios, namely Aff-Wild2~\cite{kollias2019expression}. Our primary focus is on assessing frame-level AU detection, as opposed to datasets that solely provide video-level annotations. 

\textbf{BP4D} \cite{zhang:et.al:2014:IVC}: This dataset includes 41 participants (23 females, 18 males), each appearing in 8 sessions with 2D and 3D video recordings. It contains approximately 140,000 frames, each labeled for AU occurrence and annotated with 68 facial landmarks. A three-fold cross-validation is used for evaluation.

\textbf{DISFA} \cite{mavadati:et.al:2013:TAC}: Comprising 27 videos from 15 men and 12 women, this dataset includes 130,788 frames annotated with 68 facial landmarks and AU intensities (0-5). Frames with intensities $\geq$ 2 are positive; others are negative. Three-fold cross-validation is applied for evaluation, similar to BP4D.

\textbf{BP4D+} \cite{zhang2016multimodal}: Featuring 82 females and 58 males across 10 sessions, this dataset expands upon BP4D in scale and diversity. Of the 10 sessions, 4 provide 197,875 AU-labeled frames annotated with 49 facial landmarks. Following \cite{shao2021jaa}, we train on BP4D and test on BP4D+.

\textbf{GFT} \cite{girard2017sayette}:  This dataset records 96 participants in 32 unscripted conversational groups, primarily capturing moderate out-of-plane poses. Annotations include 10 AUs and 68 facial landmarks. Training involves 78 participants (108,000 frames), while testing uses 18 participants (24,600 frames) based on standard splits.

\textbf{Aff-Wild2} \cite{kollias2019expression}: This in-the-wild dataset includes 305 training videos (1.39 million frames) and 105 validation videos (440,000 frames), sourced from YouTube with diverse demographics and conditions. Each frame is annotated with 12 AUs and 68 facial landmarks. We follow \cite{shao2023facial} by training on the training set and evaluating on the validation set.

The occurrence rates of AUs in the training sets of these datasets are presented in Table~\ref{tab:1}. It is evident from the table that the class imbalance issue of each type of AU is prevalent across all datasets, especially on the DISFA, GFT and Aff-Wild2 datasets.

\begin{table}[t!]
	\begin{center}
		\scriptsize
		\caption{AU occurrence rates (\%) in the training sets of different datasets. "-" denotes the AU is not annotated in this dataset.}
		\label{tab:1}
		\renewcommand{\arraystretch}{0.8}
		\tabcolsep=0.13cm
		\begin{tabular}{c c c c c c c c c c c c c c c c}
			\toprule[2pt]
			\multicolumn{1}{c}{\multirow{2}{*}{$\textbf{Dataset}$}}&\multicolumn{15}{c}{$\textbf{AU Index}$}\\ 
			\cline{2-16}
			\multicolumn{1}{c}{}&$\textbf{1}$&$\textbf{2}$&$\textbf{4}$&$\textbf{6}$&$\textbf{7}$&$\textbf{9}$&$\textbf{10}$&$\textbf{12}$&$\textbf{14}$&$\textbf{15}$&$\textbf{17}$&$\textbf{23}$&$\textbf{24}$&$\textbf{25}$&$\textbf{26}$ \\
			\midrule[1pt]
			\multicolumn{1}{c}{BP4D~\cite{zhang:et.al:2014:IVC}}&21.1&17.1&20.3&46.2&54.9&-&59.4&56.2&46.6&16.9&34.4&16.5&15.2&-&-\\ 
			\multicolumn{1}{c}{BP4D+~\cite{zhang2016multimodal}}&10.0&8.0&6.0&50.0&66.0&-&65.0&58.0&60.0&11.0&13.0&17.0&4.0&-&-\\
			\multicolumn{1}{c}{DISFA~\cite{mavadati:et.al:2013:TAC}}&5.0&4.0&15.0&8.1&-&4.3&-&13.2&-&-&-&-&-&27.8&8.9\\  
			\multicolumn{1}{c}{GFT~\cite{girard2017sayette}}&3.7&13.5&3.7&28.3&-&-&24.6&29.3&3.1&10.7&-&25.0&14.1&-&-\\ 	
			\multicolumn{1}{c}{Aff-Wild2~\cite{kollias2019expression}}&11.9&5.1&16.0&26.5&39.9&-&34.5&24.3&-&2.8&-&3.1&2.8&62.8&7.6\\ 	
			\bottomrule[2pt]
		\end{tabular}
	\end{center}
\end{table}

\subsubsection{Implementation Details} 
Following the preprocessing method used in MEGraph~\cite{luo2022learning}, the 68 facial landmarks undergo a conversion into 49 facial internal landmarks, excluding those associated with the facial contour. Subsequently, the images undergo augmentation through similarity transformation, encompassing in-plane rotation, uniform scaling, and translation, utilizing the 49 facial landmarks. The resulting images are standardized to a resolution of 256$\times$256$\times$3 pixels and then randomly cropped into 224$\times$224$\times$3 pixels, with the addition of a random horizontal flip.

For the pre-training process, the encoder is conducted on one NVIDIA GeForce RTX 3090 GPU with 24GB for one day and implemented by PyTorch using an AdamW optimizer with $\beta_{1}$= 0.9, $\beta_{2}$= 0.999 and weight decay of $10^{-6}$. We set a learning rate of 10$^{-5}$ and 60 epochs. In our discriminative contrastive learning framework, where sample pairs are selected within the same batch, the batch size significantly impacts the experimental results. For consistency, we have set the batch size to 32 in our experiments. In terms of the hyperparameters in AUNCE, $n_{au}$ is set as 12 for BP4D, BP4D+, Aff-Wild2, 10 for GFT, and 8 for DISFA respectively, and the temperature coefficient $\tau$ is 0.5. The analysis for $\beta$ and $p_{1}$, $p_{2}$, $p_{3}$ will introduce in Section.~\ref{beta}. 

For the linear evaluation protocol~\cite{henaff2020data,chen2020simple}, we extract $n_{au}$ 2048-dimensional feature vectors from the model pre-trained using the AUNCE loss. A linear classifier is then trained on these features using AdamW optimizer with the same settings as the pre-training phase, a batch size of 256 on a single GPU, a learning rate of 10$^{-3}$, and 100 epochs. The results are obtained through training with the weighted cross-entropy loss (WCE).

In addition, we report the standard deviations of the experimental results in Tables 3 through 7. These deviations are computed over at least five independent runs to demonstrate the robustness of the proposed framework.

\subsubsection{Evaluation Metrics}
Aligned with current state-of-the-art methodologies, we evaluate our AU detection method using three frame-based metrics: F1-macro score, F1-micro score, and Accuracy. The F1-macro score is obtained by calculating the F1 score for each AU class individually and then averaging these values, which is a standard metric employed by many existing methods. Inspired by the work of Hinduja \textit{et al.} \cite{hinduja2024time}, we also incorporate the F1-micro score, which is computed by aggregating the true positives, false positives, and false negatives across all classes before calculating the overall F1-score. This metric is particularly useful for evaluating performance in imbalanced datasets. Additionally, we include Accuracy, which is defined as the ratio of correctly classified frames to the total number of frames, to provide a more comprehensive assessment of model performance. For simplicity, we will refer to the F1-macro score as F1-score in the remainder of the text.

\begin{table}[t!]
	\begin{center}
		\scriptsize
		\caption{Quantitative comparison results on BP4D dataset, where the evaluation metric is F1-score(\%).}
		\label{tab:2}
		\renewcommand{\arraystretch}{0.8}
		\tabcolsep=0.07cm
		\begin{tabular}{l l c c c c c c c c c c c c c}
			\toprule[1.5pt]
			\multicolumn{1}{c}{\multirow{2}{*}{$\textbf{Method}$}}&\multicolumn{1}{c}{\multirow{2}{*}{$\textbf{Source}$}}&\multicolumn{12}{c}{$\textbf{AU Index}$}&\multicolumn{1}{c}{\multirow{2}{*}{$\textbf{Avg.}$}}\\ 
			\cline{3-14}
			\multicolumn{1}{l}{}&&$\textbf{1}$&$\textbf{2}$&$\textbf{4}$&$\textbf{6}$&$\textbf{7}$&$\textbf{10}$&$\textbf{12}$&$\textbf{14}$&$\textbf{15}$&$\textbf{17}$&$\textbf{23}$&$\textbf{24}$& \\
			\midrule[1pt]
			\multicolumn{15}{l}{Multi-task learning-based methods}\\
			\hline
			\multicolumn{1}{l}{MAL~\cite{li2023meta}}&2023 TAC&47.9&49.5&52.1&77.6&77.8&82.8&88.3&66.4&49.7&59.7&45.2&48.5&62.2\\
			\multicolumn{1}{l}{J$\hat{\rm A}$A-Net~\cite{shao2021jaa}}&2021 IJCV&53.8&47.8&58.2&78.5&75.8&82.7&88.2&63.7&43.3&61.8&45.6&49.9&62.4\\ 	
			\multicolumn{1}{l}{GeoConv~\cite{chen2022geoconv}}&2022 PR&48.4&44.2&59.9&78.4&75.6&83.6&86.7&65.0&53.0&64.7&49.5&54.1&63.6\\
			\hline
			\multicolumn{15}{l}{Semantic prior knowledge-based methods}\\
			\hline
			\multicolumn{1}{l}{MMA-Net~\cite{shang2023mma}}&2023 PRL&52.5&50.9&58.3&76.3&75.7&83.8&87.9&63.8&48.7&61.7&46.5&54.4&63.4\\
			\multicolumn{1}{l}{SEV-Net~\cite{yang2021exploiting}}&2021 CVPR&$\textbf{58.2}$&50.4&58.3&$\textbf{81.9}$&73.9&$\textbf{87.7}$&87.5&61.6&52.6&62.2&44.6&47.6&63.9\\
			\multicolumn{1}{l}{ETDF~\cite{chang2025facial}}&2025 TAC&54.7&50.8&57.1&78.8&79.6&84.6&88.0&67.0&54.9&62.9&48.6&54.5&65.1\\
			\hline
			\multicolumn{15}{l}{Attention mechanism-based methods}\\
			\hline
			\multicolumn{1}{l}{ARL~\cite{shao2022facial}}&2022 TAC&45.8&39.8&55.1&75.7&77.2&82.3&86.6&58.8&47.6&62.1&47.4&55.4&61.1\\
			\multicolumn{1}{l}{AC$^{2}$D~\cite{shao2025facial}}&2025 IJCV&54.2&$\textbf{54.7}$&56.5&77.0&76.2&84.0&89.0&63.6&54.8&63.6&46.5&54.8&64.6\\
			\multicolumn{1}{l}{ESCDA~\cite{shao2026constrained}}&2026 PR&56.5&44.5&59.6&79.4&77.4&84.7&88.7&64.9&$\textbf{57.0}$&65.0&52.5&55.6&65.5\\
			\hline 
			\multicolumn{15}{l}{Correlational information-based methods}\\
			\hline
			\multicolumn{1}{l}{AAR~\cite{shao2023facial}}&2023 TIP&53.2&47.7&56.7&75.9&79.1&82.9&88.6&60.5&51.5&61.9&51.0&56.8&63.8\\   
			\multicolumn{1}{l}{MEGraph~\cite{luo2022learning}}&2022 IJCAI&52.7&44.3&60.9&79.9&80.1&85.3&89.2&$\textbf{69.4}$&55.4&64.4&49.8&55.1&65.5\\
			\multicolumn{1}{l}{SACL~\cite{liu2025multi}}&2025 TAC&57.8&48.8&59.4&79.1&78.8&84.0&88.2&65.2&56.1&63.8&50.8&55.2&65.6\\ 
			\hline
			\multicolumn{15}{l}{Contrastive learning-based methods}\\
			\hline
			\multicolumn{1}{l}{SimCLR~\cite{chen2020simple}}&2020 ICML&38.0&36.4&37.2&66.6&64.7&76.2&76.2&51.1&29.8&56.1&27.5&37.7&49.8\\  
			\multicolumn{1}{l}{MoCo~\cite{he2020momentum}}&2020 CVPR&30.8&41.3&42.1&70.2&70.4&78.7&82.5&53.3&25.2&59.1&31.5&34.3&51.6\\ 
			\multicolumn{1}{l}{EmoCo~\cite{sun2021emotion}}&2021 FG&50.2&44.7&53.9&74.8&76.6&83.7&87.9&61.7&47.6&59.8&46.9&54.6&61.9\\  
			\multicolumn{1}{l}{CLP~\cite{li2024contrastive}}&2024 TIP&47.7&50.9&49.5&75.8&78.7&80.2&84.1&67.1&52.0&62.7&45.7&54.8&62.4\\ 
			\multicolumn{1}{l}{KSRL~\cite{chang2022knowledge}}&2022 CVPR&53.3&47.4&56.2&79.4&$\textbf{80.7}$&85.1&89.0&67.4&55.9&61.9&48.5&49.0&64.5\\   
			\multicolumn{1}{l}{CLEF~\cite{zhang2023weakly}}&2023 ICCV&55.8&46.8&$\textbf{63.3}$&79.5&77.6&83.6&87.8&67.3&55.2&63.5&$\textbf{53.0}$&57.8&65.9\\  
			\multicolumn{1}{l}{$\textbf{AUNCE(Ours)}$}&\multicolumn{1}{c}{-}&53.6&49.8&61.6&78.4&78.8&84.7&$\textbf{89.6}$&67.4&55.1&$\textbf{65.4}$&50.9&$\textbf{58.0}$&$\textbf{66.1}$\\ 
			\multicolumn{1}{l}{}&\multicolumn{1}{l}{}&($\pm$&($\pm$&($\pm$&($\pm$&($\pm$&($\pm$&($\pm$&($\pm$&($\pm$&($\pm$&($\pm$&($\pm$&($\pm$\\ 
			\multicolumn{1}{l}{}&\multicolumn{1}{l}{}&0.15)&0.12)&0.17)&0.21)&0.19)&0.22)&0.22)&0.17)&0.14)&0.15)&0.14)&0.15)&0.17)\\ 
			\bottomrule[1.5pt]
		\end{tabular}
	\end{center}
\end{table}

\begin{table}[t!]
	\begin{center}
		\scriptsize
		\caption{Quantitative comparison results on DISFA dataset, where the evaluation metric is F1-score(\%).}
		\label{tab:3}
		\renewcommand{\arraystretch}{0.8}
		\tabcolsep=0.19cm
		\begin{tabular}{l l c c c c c c c c c}
			\toprule[2pt]
			\multicolumn{1}{c}{\multirow{2}{*}{$\textbf{Method}$}}&\multicolumn{1}{c}{\multirow{2}{*}{$\textbf{Source}$}}&\multicolumn{8}{c}{$\textbf{AU Index}$}&\multicolumn{1}{c}{\multirow{2}{*}{$\textbf{Avg.}$}}\\ 
			\cline{3-10}
			\multicolumn{1}{l}{}&&$\textbf{1}$&$\textbf{2}$&$\textbf{4}$&$\textbf{6}$&$\textbf{9}$&$\textbf{12}$&$\textbf{25}$&$\textbf{26}$& \\
			\midrule[1pt]
			\multicolumn{11}{l}{Multi-task learning-based methods}\\
			\hline
			\multicolumn{1}{l}{MAL~\cite{li2023meta}}&2023 TAC&43.8&39.3&68.9&47.4&48.6&72.7&90.6&52.6&58.0\\
			\multicolumn{1}{l}{GeoConv~\cite{chen2022geoconv}}&2022 PR&$\textbf{65.5}$&$\textbf{65.8}$&67.2&48.6&51.4&72.6&80.9&44.9&62.1\\
			\multicolumn{1}{l}{J$\hat{\rm A}$A-Net~\cite{shao2021jaa}}&2021 IJCV&62.4&60.7&67.1&41.1&45.1&73.5&90.9&67.4&63.5\\
			\hline
			\multicolumn{11}{l}{Semantic prior knowledge-based methods}\\
			\hline
			\multicolumn{1}{l}{SEV-Net~\cite{yang2021exploiting}}&2021 CVPR&55.3&53.1&61.5&53.6&38.2&71.6&$\textbf{95.7}$&41.5&58.8\\
			\multicolumn{1}{l}{ETDF~\cite{chang2025facial}}&2025 TAC&62.6&54.7&70.8&46.3&51.7&76.3&94.4&59.8&64.6\\
			\multicolumn{1}{l}{GLEE-Net~\cite{zhang2024detecting}}&2024 TCSVT&61.9&54.0&75.8&45.9&55.7&77.6&92.9&60.0&65.5\\  
			\hline
			\multicolumn{11}{l}{Attention mechanism-based methods}\\
			\hline
			\multicolumn{1}{l}{ROI~\cite{li2017action}}&2017 CVPR&41.5&26.4&66.4&50.7&$\textbf{80.5}$&$\textbf{89.3}$&88.9&15.6&48.5\\
			\multicolumn{1}{l}{ARL~\cite{shao2022facial}}&2022 TAC&43.9&42.1&63.6&41.8&40.0&76.2&95.2&66.8&58.7\\
			\multicolumn{1}{l}{AC$^{2}$D~\cite{shao2025facial}}&2025 IJCV&57.8&59.2&70.1&50.1&54.4&75.1&90.3&66.2&65.4\\
			\hline 
			\multicolumn{11}{l}{Correlational information-based methods}\\
			\hline  
			\multicolumn{1}{l}{MEGraph~\cite{luo2022learning}}&2022 IJCAI&52.5&45.7&$\textbf{76.1}$&51.8&46.5&76.1&92.9&57.6&62.4\\			
			\multicolumn{1}{l}{AAR~\cite{shao2023facial}}&2023 TIP&62.4&53.6&71.5&39.0&48.8&76.1&91.3&$\textbf{70.6}$&64.2\\
			\multicolumn{1}{l}{SACL~\cite{liu2025multi}}&2025 TAC&62.0&65.7&74.5&53.2&43.1&76.9&95.6&53.1&65.5\\
			\hline
			\multicolumn{11}{l}{Contrastive learning-based methods}\\
			\hline
			\multicolumn{1}{l}{SIMCLR~\cite{chen2020simple}}&2020 ICML&21.2&23.3&47.5&42.4&35.5&66.8&81.5&52.7&46.4\\
			\multicolumn{1}{l}{MoCo~\cite{he2020momentum}}&2020 CVPR&22.7&18.2&45.9&45.4&34.1&72.9&83.4&54.5&47.1\\  
			\multicolumn{1}{l}{CLP~\cite{li2024contrastive}}&2024 TIP&42.4&38.7&63.5&$\textbf{59.7}$&38.9&73.0&85.0&58.1&57.4\\
			\multicolumn{1}{l}{EmoCo~\cite{sun2021emotion}}&2021 FG&42.7&41.0&66.3&45.1&50.9&75.5&88.9&58.6&58.6\\
			\multicolumn{1}{l}{KSRL~\cite{chang2022knowledge}}&2022 CVPR&60.4&59.2&67.5&52.7&51.5&76.1&91.3&57.7&64.5\\ 		
			\multicolumn{1}{l}{CLEF~\cite{zhang2023weakly}}&2023 ICCV&64.3&61.8&68.4&49.0&55.2&72.9&89.9&57.0&64.8\\
			\multicolumn{1}{l}{$\textbf{AUNCE(Ours)}$}&\multicolumn{1}{c}{-}&61.8&58.9&74.9&49.7&56.2&73.5&92.1&64.2&$\textbf{66.4}$\\ 
			\multicolumn{1}{l}{}&\multicolumn{1}{l}{}&($\pm$&($\pm$&($\pm$&($\pm$&($\pm$&($\pm$&($\pm$&($\pm$&($\pm$\\
			\multicolumn{1}{l}{}&\multicolumn{1}{l}{}&0.37)&0.22)&0.42)&0.31)&0.36)&0.49)&0.42)&0.28)&0.36)\\
			\bottomrule[2pt] 
		\end{tabular}
	\end{center}
\end{table}

\begin{table}[t!] 
	\begin{center}
		\scriptsize
		\caption{Quantitative comparison results on BP4D+ dataset in terms of cross-dataset evaluation, where the evaluation metric is F1-score(\%).}
		\label{tab:4}
		\renewcommand{\arraystretch}{0.8}
		\tabcolsep=0.07cm
		\begin{tabular}{l l c c c c c c c c c c c c c}
			\toprule[2pt]
			\multicolumn{1}{c}{\multirow{2}{*}{$\textbf{Method}$}}&\multicolumn{1}{c}{\multirow{2}{*}{$\textbf{Source}$}}&\multicolumn{12}{c}{$\textbf{AU Index}$}&\multicolumn{1}{c}{\multirow{2}{*}{$\textbf{Avg.}$}}\\ 
			\cline{3-14}
			\multicolumn{1}{l}{}&&$\textbf{1}$&$\textbf{2}$&$\textbf{4}$&$\textbf{6}$&$\textbf{7}$&$\textbf{10}$&$\textbf{12}$&$\textbf{14}$&$\textbf{15}$&$\textbf{17}$&$\textbf{23}$&$\textbf{24}$& \\
			\midrule[1pt]
			\multicolumn{1}{l}{ARL~\cite{shao2022facial}}&2022 TAC&29.9&33.1&27.1&81.5&83.0&84.8&86.2&59.7&44.6&43.7&48.8&32.3&54.6\\
			\multicolumn{1}{l}{J$\hat{\rm A}$A-Net~\cite{shao2021jaa}}&2021 IJCV&39.7&35.6&30.7&82.4&84.7&88.8&87.0&62.2&38.9&46.4&48.9&$\textbf{36.0}$&56.8\\
			\multicolumn{1}{l}{AC$^{2}$D~\cite{shao2025facial}}&2025 IJCV&42.3&35.4&26.7&80.7&87.0&90.9&85.8&73.3&45.3&43.4&50.3&29.0&57.5\\
			\multicolumn{1}{l}{GLEE-Net~\cite{zhang2024detecting}}&2024 TCSVT&39.8&37.9&$\textbf{41.6}$&83.4&88.2&90.2&87.4&$\textbf{76.6}$&48.3&42.9&47.7&29.8&59.5\\
			\multicolumn{1}{l}{$\textbf{AUNCE(Ours)}$}&\multicolumn{1}{c}{-}&$\textbf{45.5}$&$\textbf{44.5}$&30.9&$\textbf{87.1}$&$\textbf{89.5}$&$\textbf{94.7}$&$\textbf{94.3}$&72.1&$\textbf{52.7}$&$\textbf{51.8}$&$\textbf{52.0}$&35.5&$\textbf{62.8}$\\ 
			\multicolumn{1}{l}{}&\multicolumn{1}{l}{}&($\pm$&($\pm$&($\pm$&($\pm$&($\pm$&($\pm$&($\pm$&($\pm$&($\pm$&($\pm$&($\pm$&($\pm$&($\pm$\\
			\multicolumn{1}{l}{}&\multicolumn{1}{l}{}&0.18)&0.46)&0.14)&0.32)&0.23)&0.17)&0.14)&0.49)&0.33)&0.25)&0.12)&0.24)&0.26)\\
			\bottomrule[2pt] 
		\end{tabular}
	\end{center}
\end{table}

\begin{table}[t!] 
	\begin{center}
		\scriptsize
		\caption{Quantitative comparison results on GFT dataset, where the evaluation metric is F1-score(\%).}
		\label{tab:5}
		\renewcommand{\arraystretch}{0.8}
		\tabcolsep=0.12cm
		\begin{tabular}{l l c c c c c c c c c c c}
			\toprule[2pt]
			\multicolumn{1}{c}{\multirow{2}{*}{$\textbf{Method}$}}&\multicolumn{1}{c}{\multirow{2}{*}{$\textbf{Source}$}}&\multicolumn{10}{c}{$\textbf{AU Index}$}&\multicolumn{1}{c}{\multirow{2}{*}{$\textbf{Avg.}$}}\\ 
			\cline{3-12}
			\multicolumn{1}{l}{}&&$\textbf{1}$&$\textbf{2}$&$\textbf{4}$&$\textbf{6}$&$\textbf{10}$&$\textbf{12}$&$\textbf{14}$&$\textbf{15}$&$\textbf{23}$&$\textbf{24}$& \\
			\midrule[1pt]
			\multicolumn{1}{l}{MoCo~\cite{he2020momentum}}&2020 CVPR&35.9&45.4&13.5&$\textbf{83.4}$&71.3&78.1&23.3&37.3&26.6&50.7&46.5\\
			\multicolumn{1}{l}{SimCLR~\cite{chen2020simple}}&2020 ICML&39.6&48.3&5.6&80.7&76.2&80.6&18.1&41.6&46.1&43.8&48.1\\
			\multicolumn{1}{l}{ARL~\cite{shao2022facial}}&2022 TAC&51.9&45.9&13.7&79.2&75.5&82.8&0.1&44.9&59.2&47.5&50.1\\
			\multicolumn{1}{l}{J$\hat{\rm A}$A-Net~\cite{shao2021jaa}}&2021 IJCV&46.5&49.3&19.2&79.0&75.0&84.8&44.1&33.5&54.9&50.7&53.7\\
			\multicolumn{1}{l}{CLP~\cite{li2024contrastive}}&2024 TIP&44.6&58.7&34.7&75.9&78.6&86.6&20.3&44.8&56.4&42.2&54.3\\
			\multicolumn{1}{l}{AAR~\cite{shao2023facial}}&2023 TIP&$\textbf{66.3}$&53.9&23.7&81.5&73.6&84.2&43.8&$\textbf{53.8}$&58.2&46.5&58.5\\
			\multicolumn{1}{l}{MAL~\cite{li2023meta}}&2023 TAC&52.4&57.0&$\textbf{54.1}$&74.5&78.0&84.9&43.1&47.7&54.4&51.9&59.8\\     
			\multicolumn{1}{l}{AC$^{2}$D~\cite{shao2025facial}}&2025 IJCV&60.9&58.2&24.4&83.3&75.9&87.4&$\textbf{56.4}$&46.5&$\textbf{58.3}$&50.9&60.2\\  
			\multicolumn{1}{l}{$\textbf{AUNCE(Ours)}$}&\multicolumn{1}{c}{-}&53.6&$\textbf{62.8}$&51.3&80.6&$\textbf{82.1}$&$\textbf{88.2}$&48.2&49.8&54.6&$\textbf{58.9}$&$\textbf{63.0}$\\ 
			\multicolumn{1}{l}{}&\multicolumn{1}{l}{}&($\pm$&($\pm$&($\pm$&($\pm$&($\pm$&($\pm$&($\pm$&($\pm$&($\pm$&($\pm$&($\pm$\\
			\multicolumn{1}{l}{}&\multicolumn{1}{l}{}&0.33)&0.18)&0.28)&0.24)&0.41)&0.11)&0.35)&0.23)&0.42)&0.35)&0.29)\\
			\bottomrule[2pt]  
		\end{tabular}
	\end{center}
\end{table}

\begin{table}[t!] 
	\begin{center}
		\scriptsize
		\caption{Quantitative comparison results on Aff-Wild2 dataset, where the evaluation metric is F1-score(\%).}
		\label{tab:6}
		\renewcommand{\arraystretch}{0.8}
		\tabcolsep=0.07cm
		\begin{tabular}{l l c c c c c c c c c c c c c}
			\toprule[2pt]
			\multicolumn{1}{c}{\multirow{2}{*}{$\textbf{Method}$}}&\multicolumn{1}{c}{\multirow{2}{*}{$\textbf{Source}$}}&\multicolumn{12}{c}{$\textbf{AU Index}$}&\multicolumn{1}{c}{\multirow{2}{*}{$\textbf{Avg.}$}}\\ 
			\cline{3-14}
			\multicolumn{1}{l}{}&&$\textbf{1}$&$\textbf{2}$&$\textbf{4}$&$\textbf{6}$&$\textbf{7}$&$\textbf{10}$&$\textbf{12}$&$\textbf{15}$&$\textbf{23}$&$\textbf{24}$&$\textbf{25}$&$\textbf{26}$& \\
			\midrule[1pt]
			\multicolumn{1}{l}{ARL~\cite{shao2022facial}}&2022 TAC&59.2&48.2&54.9&70.0&83.4&80.3&72.0&0.1&0.1&17.3&93.0&37.5&51.3\\
			\multicolumn{1}{l}{J$\hat{\rm A}$A-Net~\cite{shao2021jaa}}&2021 IJCV&61.7&50.1&56.0&71.7&81.7&82.3&78.0&$\textbf{31.1}$&1.4&8.6&94.8&37.5&54.6\\
			\multicolumn{1}{l}{AAR~\cite{shao2023facial}}&2023 TIP&65.4&57.9&59.9&73.2&84.6&$\textbf{83.2}$&79.9&21.8&27.4&19.9&94.5&41.7&59.1\\
			\multicolumn{1}{l}{AC$^{2}$D~\cite{shao2025facial}}&2025 IJCV&63.8&53.1&$\textbf{66.0}$&66.6&80.7&80.1&78.0&30.3&26.5&$\textbf{29.2}$&93.3&41.4&59.1\\  
			\multicolumn{1}{l}{$\textbf{AUNCE(Ours)}$}&\multicolumn{1}{c}{-}&$\textbf{68.8}$&$\textbf{67.6}$&64.7&$\textbf{74.8}$&$\textbf{88.6}$&78.4&$\textbf{81.3}$&26.9&$\textbf{36.7}$&18.6&$\textbf{96.1}$&$\textbf{43.8}$&$\textbf{62.2}$\\ 
			\multicolumn{1}{l}{}&\multicolumn{1}{l}{}&($\pm$&($\pm$&($\pm$&($\pm$&($\pm$&($\pm$&($\pm$&($\pm$&($\pm$&($\pm$&($\pm$&($\pm$&($\pm$\\
			\multicolumn{1}{l}{}&\multicolumn{1}{l}{}&0.25)&0.37)&0.28)&0.33)&0.12)&0.47)&0.31)&0.17)&0.08)&0.42)&0.39)&0.32)&0.29)\\
			\bottomrule[2pt] 
		\end{tabular}
	\end{center}
\end{table}

\begin{table}[t!] 
	\begin{center}
		\scriptsize
		\caption{Quantitative comparison results for common 10 AUs of cross evaluation between BP4D and GFT. BP4D → GFT denotes training on BP4D and testing on GFT. The evaluation metric is F1-score(\%).}
		\label{tab:7}
		\renewcommand{\arraystretch}{0.8}
		\tabcolsep=0.1cm
		\begin{tabular}{l l l c c c c c c c c c c c}
			\toprule[2pt]
			\multicolumn{1}{c}{\multirow{2}{*}{$\textbf{Dataset}$}}&\multicolumn{1}{c}{\multirow{2}{*}{$\textbf{Method}$}}&\multicolumn{1}{c}{\multirow{2}{*}{$\textbf{Source}$}}&\multicolumn{10}{c}{$\textbf{AU Index}$}&\multicolumn{1}{c}{\multirow{2}{*}{$\textbf{Avg.}$}}\\ 
			\cline{4-13}
			\multicolumn{1}{l}{}&&&$\textbf{1}$&$\textbf{2}$&$\textbf{4}$&$\textbf{6}$&$\textbf{10}$&$\textbf{12}$&$\textbf{14}$&$\textbf{15}$&$\textbf{23}$&$\textbf{24}$&\\
			\midrule[1pt]
			\multicolumn{1}{l}{\multirow{4}{*}{BP4D → GFT}}&D$^{2}$CA~\cite{li2025decoupled}&2025 TIP&14.3&20.4&10.1&66.8&57.0&65.1&7.6&24.9&41.4&$\textbf{45.7}$&35.3\\
			\multicolumn{1}{l}{}&AC$^{2}$D~\cite{shao2025facial}&2025 IJCV&28.0&35.7&22.7&70.5&69.2&65.2&16.3&$\textbf{29.0}$&40.8&45.2&42.3\\
			\multicolumn{1}{l}{}&Ours (WCE)&\multicolumn{1}{c}{-}&30.2&37.9&$\textbf{24.8}$&66.4&72.3&62.6&20.1&28.6&44.6&38.8&42.6\\
			\multicolumn{1}{l}{}&$\textbf{Ours (AUNCE)}$&\multicolumn{1}{c}{-}&$\textbf{32.4}$&$\textbf{50.1}$&10.5&$\textbf{71.0}$&$\textbf{77.6}$&$\textbf{70.2}$&$\textbf{22.3}$&26.5&$\textbf{50.9}$&40.4&$\textbf{45.2}$\\
			\hline
			\multicolumn{1}{l}{\multirow{4}{*}{GFT → BP4D}}&Ours (WCE)&\multicolumn{1}{c}{-}&49.5&45.5&26.0&20.4&52.3&36.6&17.3&18.6&50.5&54.4&37.1\\
			\multicolumn{1}{l}{}&AC$^{2}$D~\cite{shao2025facial}&2025 IJCV&51.9&49.3&25.8&24.6&50.1&40.9&15.3&20.1&47.2&58.3&38.4\\
			\multicolumn{1}{l}{}&D$^{2}$CA~\cite{li2025decoupled}&2025 TIP&31.0&33.0&$\textbf{43.2}$&$\textbf{60.5}$&53.7&$\textbf{71.1}$&21.1&$\textbf{36.5}$&36.4&44.6&43.1\\
			\multicolumn{1}{l}{}&$\textbf{Ours (AUNCE)}$&\multicolumn{1}{c}{-}&$\textbf{55.3}$&$\textbf{53.8}$&25.4&23.9&$\textbf{61.8}$&38.8&$\textbf{23.6}$&25.5&$\textbf{59.6}$&$\textbf{65.9}$&$\textbf{43.7}$\\ 
			\bottomrule[2pt] 
		\end{tabular}
	\end{center}
\end{table}

\subsection{Comparison with State-of-the-Art Methods} \label{Comparison with State-of-the-Art Methods}
\subsubsection{Evaluation on BP4D} 
Since we use AU labels to assist the process of selecting positive sample pairs, we compare our approach with five groups of supervised AU detection frameworks. For the first four categories, we select three recent or representative methods each, while the final category, contrastive learning-based approaches, is examined in greater detail. The quantitative results of state-of-the-art methods from the past five years for BP4D dataset are shown in Table~\ref{tab:2}. Compared to the latest state-of-the-art results, our method AUNCE demonstrates significantly superior performance.

Methods in the first group are based on a multi-task learning strategy, including MAL~\citep{li2023meta}, J$\hat{\rm A}$A-Net~\citep{shao2021jaa}, and GeoConv~\citep{chen2022geoconv}. Our method surpasses MAL by 3.9\%, primarily due to the ability of AUNCE to help the encoder address the class imbalance issue among different AUs, and the facial expression recognition component in MAL introduces some negative transfer effects on AU detection. The 3.7\% improvement over J$\hat{\rm A}$A-Net can be attributed to the influence of the facial landmark detection task within the multi-task learning framework on AU detection, which is also affected by the weighting of different losses. Additionally, the modest 2.5\% improvement over GeoConv is due to the relatively limited impact of integrating 3D information on enhancing the performance of 2D AU detection.

Methods in the second group mainly utilize the semantic prior knowledge, including MMA-Net~\citep{shang2023mma}, SEV-Net~\citep{yang2021exploiting} and ETDF~\citep{chang2025facial}. Our approach demonstrates an average improvement of 2.7\%, 2.2\% and 1.0\%, respectively. This phenomenon can be attributed to the inherent limitations of these methods in capturing the fine-grained and localized cues associated with individual AUs, as simple semantic descriptions often fail to reflect the underlying visual complexity. Additionally, their dependence on predefined semantic priors may introduce inductive biases that hinder the model’s ability to flexibly adapt to the nuanced and heterogeneous nature of facial behaviors across individuals and contexts.

Methods in the third group leverage attention mechanisms to selectively focus on regional facial areas surrounding AUs, including ARL~\citep{shao2022facial}, AC$^{2}$D~\citep{shao2025facial} and ESCDA~\citep{shao2026constrained}. Our approach achieves a notable improvement, with an increase in F1-score ranging from 0.6\% to 5.0\% over these methods. This advancement is primarily due to the feature extraction methods of these frameworks neglecting critical correlation information among AUs and failing to specifically address the class imbalance issue for each AU. Furthermore, their reliance on attention mechanisms may lead to suboptimal encoding when attention is misdirected or insufficiently localized, limiting the precision of AU-specific feature representations.

Methods in the fourth group focus on capturing correlational information among AUs, including AAR~\cite{shao2023facial}, SACL~\cite{liu2025multi} and MEGraph~\cite{luo2022learning}. Our approach exhibits a performance advantage of approximately 0.5\%-2.3\% in F1-score. This underscores the efficacy of our supervised discriminative contrastive learning paradigm in effectively capturing differential information between AUs from each sample pair instead of pixel-level information of the entire face associated with AU. It is noteworthy that the encoder used in our method is only the same as the network MEGraph used for the first stage of training, with reduced parameters by nearly 7M. This highlights learning only the difference information among features is more efficient than learning the whole pixel-level information.

Methods in the last group are based on contrastive learning, including SIMCLR~\cite{chen2020simple}, MoCo~\cite{he2020momentum}, EmoCo~\cite{sun2021emotion}, CLP~\cite{li2024contrastive}, KSRL~\cite{chang2022knowledge} and CLEF~\cite{zhang2023weakly}. When compared to these methods, our approach demonstrates a performance advantage of approximately 0.2\% to 16.3\% in F1-score. This improvement is attributed to the efficacy of our proposed negative sample re-weighting strategy and positive sample sampling strategy, specifically tailored for the AU detection task using static images. In addition, compared to contrastive learning methods based on weighted cross-entropy loss, our approach significantly reduces the extraction of pixel-level redundant information.

\subsubsection{Evaluation on DISFA}
Comparable outcomes are evident in the DISFA dataset, as outlined in Table~\ref{tab:3}, where AUNCE attains the highest overall performance in AU detection. Furthermore, a comparison between Table~\ref{tab:2} and Table~\ref{tab:3} reveals that recent methods such as EmoCo, CLP, and MEGraph demonstrate strong performance on the BP4D dataset but achieve only mediocre results on the DISFA dataset. This performance gap is likely attributed to the more pronounced class imbalance issue in DISFA, as illustrated in Table~\ref{tab:1}. In contrast, AUNCE consistently delivers robust performance on both BP4D and DISFA datasets and achieves more stable results across various AUs due to our proposed negative sample re-weighting strategy.

\subsubsection{Evaluation on BP4D+}
To assess the model's performance on a larger and more diverse test set, we train AUNCE on the full BP4D and evaluate it on the entire BP4D+ using a cross-dataset testing protocol. The comparative results are presented in Table~\ref{tab:4}. Notably, AUNCE achieves superior performance in terms of the average F1-score, outperforming previous methods. This indicates that AUNCE maintains robust performance even when the testing data significantly increases in both scale and diversity.

\subsubsection{Evaluation on GFT}
The F1-score results on GFT are summarized in Table~\ref{tab:5}, where the proposed AUNCE method demonstrates a clear advantage over previous approaches. Unlike BP4D, BP4D+, and DISFA, which predominantly consist of near-frontal facial images, GFT features a substantial number of images captured with out-of-plane poses, posing greater challenges for accurate AU detection. While methods such as ARL, CLP and AC$^{2}$D encounter significant performance limitations under these conditions, AUNCE achieves an impressive average F1-score of 63.0\%, highlighting its robustness in handling complex pose variations.

\subsubsection{Evaluation on Aff-Wild2}
We have validated the performance of our AUNCE method in constrained scenarios. To further assess its effectiveness in unconstrained scenarios, we compared it with other approaches on the challenging Aff-Wild2 dataset, as shown in Table~\ref{tab:6}. Our AUNCE method demonstrates a higher average F1-score than previous methods. Notably, while EAC-Net leverage external training data, our method achieves superior overall performance using only the Aff-Wild2 dataset. By comparing the results of ARL, J$\hat{\rm A}$A-Net, AAR and AC$^{2}$D to our AUNCE method across Table~\ref{tab:2} to Table~\ref{tab:6}, it is evident that the performance margins between our AUNCE method and other methods on GFT and Aff-Wild2 are larger than those on BP4D, BP4D+ and DISFA. This indicates that AUNCE is more effective at handling challenging cases in AU detection. 

\subsubsection{Cross dataset validation of AUNCE}
To assess the generalization capability and cross-dataset robustness of our framework, we compare AUNCE against three baselines: a model using the same encoder with weighted cross-entropy loss (WCE), AC$^{2}$D~\cite{shao2025facial}, and D$^{2}$CA~\cite{li2025decoupled}, the latter of which is specifically designed for cross-domain AU detection. The comparative results are summarized in Table~\ref{tab:7}. Given that 10 AUs in the GFT dataset overlap with 12 AUs in BP4D, we conduct cross-dataset evaluations between BP4D and GFT.

For the BP4D-to-GFT setting, we utilize three BP4D models trained via 3-fold cross-validation and report their averaged performance. Conversely, for the GFT-to-BP4D setting, the model trained on GFT is evaluated on the three BP4D test folds, with average results reported accordingly.

As shown in Tables~\ref{tab:2} and~\ref{tab:5}, AUNCE exhibits a performance drop in cross-dataset settings, largely due to the significant domain shift between BP4D and GFT. Nevertheless, AUNCE consistently outperforms all three baselines in both BP4D-to-GFT and GFT-to-BP4D scenarios, underscoring its superior generalization ability and robustness across domains.

\begin{table}[hbtp]
	\begin{center} 
		\scriptsize
		\caption{Comparison of the AUNCE loss with the weighted cross-entropy loss (WCE) and the weighted focal loss (WFL) during the pretraining phase across different encoders on BP4D and DISFA datasets.}
		\label{tab:comparision}
		\renewcommand{\arraystretch}{0.8}
		\tabcolsep=0.5cm
		\begin{tabular}{c c c c c}
			\toprule[2pt]
			\multicolumn{1}{c}{\multirow{2}{*}{$\textbf{Encoder}$}}&\multicolumn{1}{c}{\multirow{2}{*}{$\textbf{Source}$}}&\multicolumn{1}{c}{\multirow{2}{*}{$\textbf{Loss}$}}&\multicolumn{2}{c}{$\textbf{Dataset}$}\\ 
			\cline{4-5}
			\multicolumn{1}{l}{}&&&$\textbf{BP4D}$&$\textbf{DISFA}$\\
			\midrule[1pt]    
			\multicolumn{1}{c}{\multirow{3}{*}{ResNet50~\cite{he2016deep}}}&\multicolumn{1}{c}{\multirow{3}{*}{2016CVPR}}&WCE&56.2&45.2\\
			\multicolumn{1}{c}{}&&WFL&57.7$^{\textcolor{red}{1.5\uparrow}}$&47.0$^{\textcolor{red}{1.8\uparrow}}$\\
			\multicolumn{1}{c}{}&&AUNCE&$\textbf{59.7}^{\textcolor{red}{3.5\uparrow}}$&$\textbf{51.6}^{\textcolor{red}{6.4\uparrow}}$\\
			\hline
			\multicolumn{1}{c}{\multirow{3}{*}{ROI~\cite{li2017action}}}&\multicolumn{1}{c}{\multirow{3}{*}{2017CVPR}}&WCE&56.4&48.5\\
			\multicolumn{1}{c}{}&&WFL&57.0$^{\textcolor{red}{0.6\uparrow}}$&50.6$^{\textcolor{red}{2.1\uparrow}}$\\
			\multicolumn{1}{c}{}&&AUNCE&$\textbf{58.8}^{\textcolor{red}{2.4\uparrow}}$&$\textbf{52.8}^{\textcolor{red}{4.3\uparrow}}$\\ 
			\hline
			\multicolumn{1}{c}{\multirow{3}{*}{J$\hat{\rm A}$A-Net~\cite{shao2021jaa}}}&\multicolumn{1}{c}{\multirow{3}{*}{2021IJCV}}&WCE&62.4&63.5\\
			\multicolumn{1}{c}{}&&WFL&63.8$^{\textcolor{red}{1.4\uparrow}}$&63.6$^{\textcolor{red}{0.1\uparrow}}$\\
			\multicolumn{1}{c}{}&&AUNCE&$\textbf{64.1}^{\textcolor{red}{1.7\uparrow}}$&$\textbf{64.4}^{\textcolor{red}{0.9\uparrow}}$\\
			\hline
			\multicolumn{1}{c}{\multirow{3}{*}{MEGraph (Onestage)~\cite{luo2022learning}}}&\multicolumn{1}{c}{\multirow{3}{*}{2022IJCAI}}&WCE&63.6&60.8\\
			\multicolumn{1}{c}{}&&WFL&64.2$^{\textcolor{red}{0.6\uparrow}}$&61.8$^{\textcolor{red}{1.0\uparrow}}$\\
			\multicolumn{1}{c}{}&&AUNCE&$\textbf{66.1}^{\textcolor{red}{2.5\uparrow}}$&$\textbf{66.4}^{\textcolor{red}{5.6\uparrow}}$\\
			\bottomrule[2pt] 
		\end{tabular}
	\end{center}
\end{table}

\subsection{Ablation Study}
\subsubsection{Compared with weighted cross-entropy loss and weighted focal loss} 
The primary innovation of this paper is the introduction of AUNCE, a novel contrastive loss designed to be compatible with various AU detection encoders. To evaluate its effectiveness, we compare the proposed AUNCE loss against the traditional weighted cross-entropy loss (WCE) and the recently introduced weighted focal loss (WFL) \cite{shang2024facial} during the pretraining stage, using multiple AU detection encoders. While numerous loss functions have been proposed to address sample imbalance, WFL has emerged as a widely adopted and empirically validated approach for mitigating positive–negative imbalance in AU detection. The results, presented in Table~\ref{tab:comparision}, demonstrate that AUNCE consistently outperforms the other losses in AU detection. Notably, the MEGraph encoder achieves the highest performance, with scores of 66.1\% and 66.4\% on BP4D and DISFA datasets, respectively, when trained using AUNCE. This highlights the advantage of emphasizing the capture of distinct features from individual samples, rather than solely relying on pixel-level information of the entire face in relation to AU detection.

\begin{table}[hbtp] 
	\begin{center} 
		\scriptsize
		\caption{Comparison of the computational complexity and efficiency of the weighted cross-entropy loss (WCE), the weighted focal loss (WFL), and the proposed AUNCE loss on BP4D and DISFA dataset, where encoders of these three losses are both MEGraph with the first-stage training network.}
		\label{tab:comparision:efficiency}
		\renewcommand{\arraystretch}{0.8}
		\tabcolsep=0.5cm
		\begin{tabular}{c c c c c c}
			\toprule[2pt]
			$\textbf{Dataset}$&$\textbf{Loss}$&$\textbf{Complexity}$&$\textbf{Frame Rate(FPS)}$&$\textbf{1 Epoch Time(s)}$\\
			\midrule[1pt]   
			\multicolumn{1}{c}{\multirow{3}{*}{BP4D}}&WFL&$O(NC)$&0.0293&2959\\
			\multicolumn{1}{c}{}&WCE&$O(NC)$&0.0271&2734\\
			\multicolumn{1}{c}{}&AUNCE&$O(N+2)$&$\textbf{0.0138}$&$\textbf{1388}$\\
			\hline  
			\multicolumn{1}{c}{\multirow{3}{*}{DISFA}}&WFL&$O(NC)$&0.0284&2473\\
			\multicolumn{1}{c}{}&WCE&$O(NC)$&0.0265&2308\\
			\multicolumn{1}{c}{}&AUNCE&$O(N+2)$&$\textbf{0.0136}$&$\textbf{1189}$\\
			\bottomrule[2pt] 
		\end{tabular}
	\end{center}
\end{table}

We compare the computational complexity and efficiency of the three losses. WCE and WFL both have an asymptotic complexity of $O(NC)$, as both of them involve computing probabilities for $C$ classes and summing over all $N$ samples, and WFL introducing negligible overhead due to the power and multiplication operations from the probability margin $\gamma$. In contrast, AUNCE has a lower complexity of $O(N+2)$, as it processes one anchor point, one positive sample, and $N$ negative samples, making it approximately twice as efficient as WCE and WFL when $C=2$. This suggests that AUNCE, by focusing on differential information between AUs, is more efficient for training the encoder compared to methods that capture entire facial information.

To further validate this advantage, we evaluate computational efficiency using two metrics: frame rate and training time per epoch. As shown in Table~\ref{tab:comparision:efficiency}, under identical encoder parameters, AUNCE improves frame rate and reduces epoch training time by nearly half compared to WCE and WFL. These results confirm AUNCE's superior efficiency in training, delivering comparable or better performance with significantly reduced computational overhead.

\begin{table}[h!]
	\begin{center}
		\scriptsize
		\caption{Quantitative results of ablation study on BP4D dataset.}
		\renewcommand{\arraystretch}{0.8}
		\tabcolsep=0.045cm
		\label{tab:ablation}
		\begin{tabular}{c c c c c c c c c c c c c c c c c c c c}
			\toprule[2pt]
			\multicolumn{1}{c}{\multirow{2}{*}{$\textbf{Model}$}}&\multicolumn{1}{c}{\multirow{2}{*}{$\textbf{BL}$}}&\multicolumn{1}{c}{\multirow{2}{*}{$\textbf{w}_i$}}&\multicolumn{1}{c}{\multirow{2}{*}{$\textbf{PS}$}}&\multicolumn{1}{c}{\multirow{2}{*}{$\textbf{NS}$}}&\multicolumn{12}{c}{$\textbf{AU Index}$}&$\textbf{F1-Macro}$&$\textbf{F1-Micro}$&$\textbf{Accuracy}$\\ 
			\cline{6-17}
			\multicolumn{1}{l}{}&\multicolumn{1}{l}{}&\multicolumn{1}{l}{}&\multicolumn{1}{l}{}&\multicolumn{1}{l}{}&$\textbf{1}$&$\textbf{2}$&$\textbf{4}$&$\textbf{6}$&$\textbf{7}$&$\textbf{10}$&$\textbf{12}$&$\textbf{14}$&$\textbf{15}$&$\textbf{17}$&$\textbf{23}$&$\textbf{24}$&$\textbf{(\%)}$&$\textbf{(\%)}$&$\textbf{(\%)}$ \\
			\midrule[1pt]
			\multicolumn{1}{c}{A}&$\checkmark$&&&&39.8&44.3&58.8&76.0&77.7&76.9&88.0&64.5&49.0&59.0&43.5&43.4&60.1&57.0&79.1\\ 
			\multicolumn{1}{c}{B}&$\checkmark$&$\checkmark$&&&40.1&43.0&58.9&75.4&$\textbf{79.3}$&78.8&87.8&64.1&48.7&61.0&46.9&47.5&61.0$^{\textcolor{red}{0.9\uparrow}}$&57.2$^{\textcolor{red}{0.2\uparrow}}$&79.3$^{\textcolor{red}{0.2\uparrow}}$\\ 
			\multicolumn{1}{c}{C}&$\checkmark$&$\checkmark$&$\checkmark$&&51.1&45.6&58.3&77.5&77.0&84.5&89.0&63.7&54.6&$\textbf{65.5}$&46.8&53.3&63.8$^{\textcolor{red}{3.7\uparrow}}$&57.6$^{\textcolor{red}{0.6\uparrow}}$&79.8$^{\textcolor{red}{0.7\uparrow}}$\\ 
			\multicolumn{1}{c}{D}&$\checkmark$&$\checkmark$&&$\checkmark$&50.2&43.0&58.7&77.9&77.3&83.7&88.7&64.2&53.9&64.9&50.0&56.4&64.1$^{\textcolor{red}{4.0\uparrow}}$&57.7$^{\textcolor{red}{0.7\uparrow}}$&79.9$^{\textcolor{red}{0.8\uparrow}}$\\ 
			\multicolumn{1}{c}{E}&$\checkmark$&$\checkmark$&$\checkmark$&$\checkmark$&$\textbf{53.6}$&$\textbf{49.8}$&$\textbf{61.6}$&$\textbf{78.4}$&78.8&$\textbf{84.7}$&$\textbf{89.6}$&$\textbf{67.4}$&$\textbf{55.1}$&65.4&$\textbf{50.9}$&$\textbf{58.0}$&$\textbf{66.1}^{\textcolor{red}{6.0\uparrow}}$&$\textbf{58.2}^{\textcolor{red}{1.2\uparrow}}$&$\textbf{80.4}^{\textcolor{red}{1.3\uparrow}}$\\ 
			\bottomrule[2pt]
		\end{tabular}
	\end{center}
\end{table}

\subsubsection{Evaluating other components}
We further assess the effectiveness of all proposed components. Quantitative results of ablation study are summarized in Table~\ref{tab:ablation}. The baseline (BL) is based on SIMCLR~\cite{chen2020simple}, with positive sample pairs replaced by images most similar to the original ones. Other components include the weight coefficient ($w_{i}$) for both WCE and AUNCE, the positive sample sampling strategy (PS), and the negative sample re-weighting strategy (NS).

$\textbf{Weight coefficient $w_{i}$:}$ The weights $w_{i}$ have been pointed out as a fixed value in Section.~\ref{3-1}, which achieve an increase of 0.9\% in F1-macro score, 0.2\% in F1-micro score and 0.2\% in Accuracy. For an AU, its activation indicator is a binary variable, and the corresponding information content can be measured using entropy: $-\log(p_i)$. Therefore, AUs with lower activation probabilities carry more information and should be assigned higher weights. This forms the theoretical basis for our weighting strategy.

$\textbf{Positive sample sampling strategy:}$ As shown in Table~\ref{tab:ablation}, the positive sample sampling strategy improves the F1-macro score by 2.8\%, the F1-micro score by 0.4\%, and Accuracy by 0.5\% compared to model B. This demonstrates that selecting diverse positive sample pairs with appropriate probabilities helps mitigate issues related to noisy or false AU labels. The strategy utilizes three types of positive sample pairs: self-supervised signal pairs, highest similarity pairs, and mixed feature pairs. By incorporating these varied pairs, the model benefits from more informative inputs, enhancing its generalization ability and discriminative power. These improvements enable more effective AU detection, even in the presence of label noise and class imbalance, reinforcing the model's robustness across various AU detection scenarios.

\begin{figure}[t!]
	\begin{center}
		\setlength{\abovecaptionskip}{0.1cm} 
		\begin{minipage}{0.99\columnwidth}
			\centering
			\centerline{
				\label{fig:4-a}
				\includegraphics[width=\columnwidth]{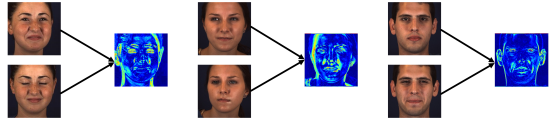}}
			\centerline{(a) Visualizing feature maps generated by model E (With re-weighting strategy)}
		\end{minipage}
		\quad   
		\begin{minipage}{0.99\columnwidth}
			\centerline{
				\label{fig:4-b}
				\includegraphics[width=\columnwidth]{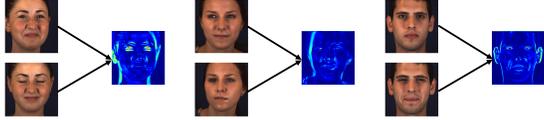}}
			\centerline{(b) Visualizing feature maps generated by model C (Without re-weighting strategy)}
		\end{minipage}
		\caption{Visualizing feature maps generated by models with and without the negative sample re-weighting strategy on BP4D dataset, which can help illustrate how the model with the negative sample re-weighting strategy concentrates more on the related facial regions. (a): Visualizing feature maps generated by model E (With the negative sample re-weighting strategy), and (b): Visualizing feature maps generated by model C (Without the negative sample re-weighting strategy).} 
		\label{fig:4}
	\end{center}
\end{figure}

$\textbf{Negative sample re-weighting strategy:}$ The negative sample re-weighting strategy, which emphasizes minority AUs during training, improves the F1-macro score by 3.1\%, the F1-micro score by 0.5\%, and overall accuracy by 0.6\% compared to model B. This demonstrates its effectiveness in adjusting backpropagation gradients between majority and minority classes, allowing more efficient learning from limited samples. Furthermore, model E shows a 2.3\% average improvement in F1-macro score over model C, with notable gains in several highly imbalanced AUs, including AU1, AU2, AU4, AU15, AU17, AU23, and AU24, further validating the strategy’s impact. To qualitatively assess its effectiveness, we additionally provide visualization results in Fig.~\ref{fig:4}. As shown, model E (top row) attends more closely to facial regions associated with rare AUs, enhancing detection performance in these difficult cases and reinforcing the method’s capacity to mitigate class imbalance.\\

\begin{figure}[t!]
	\begin{center}
		\begin{minipage}{0.16\columnwidth}
			\centering
			\centerline{
				\label{fig:5-a}
				\includegraphics[width=\columnwidth]{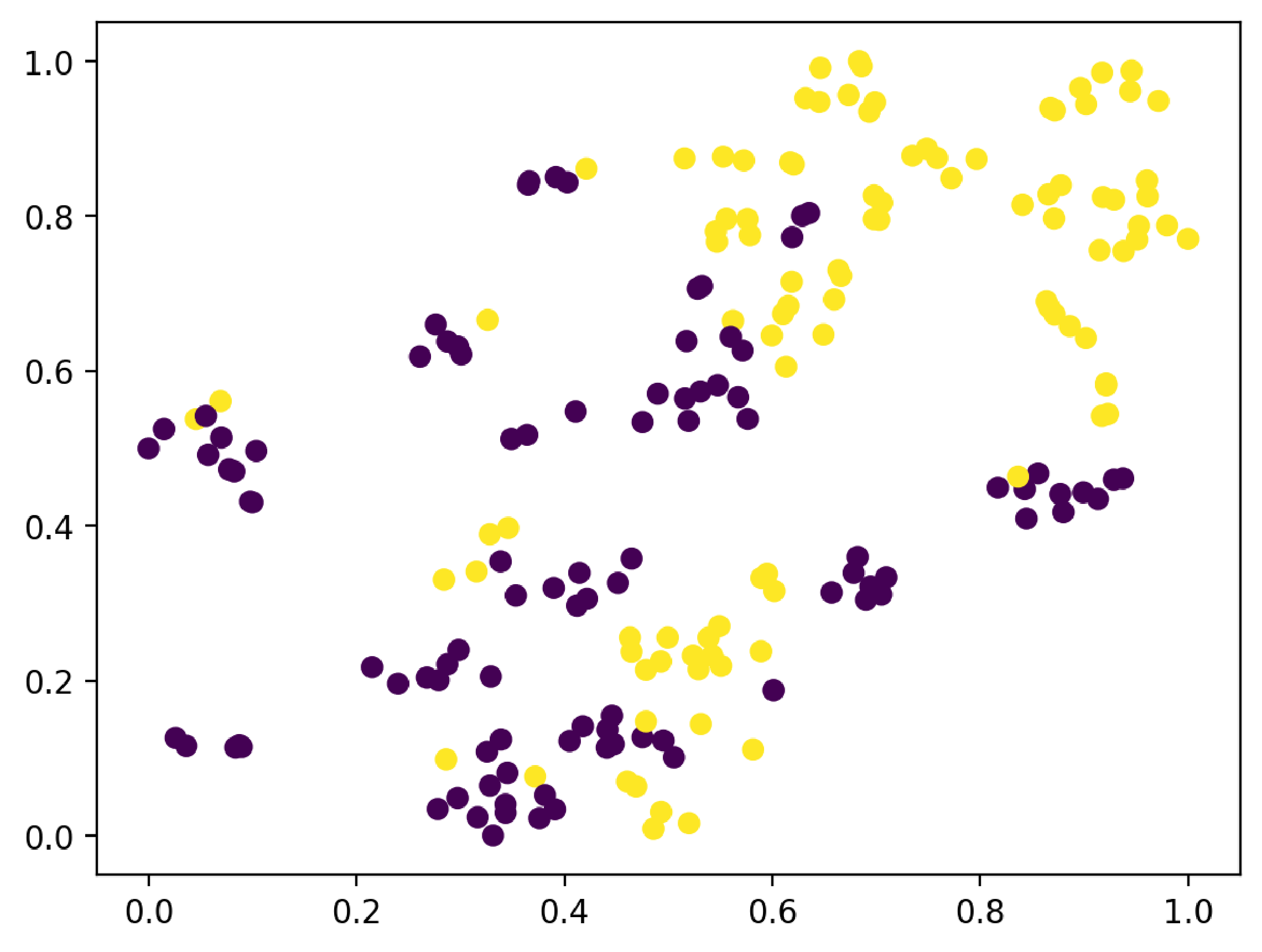}}
			\centerline{(a)}
		\end{minipage}
		\begin{minipage}{0.16\columnwidth}
			\centerline{
				\label{fig:5-b}
				\includegraphics[width=\columnwidth]{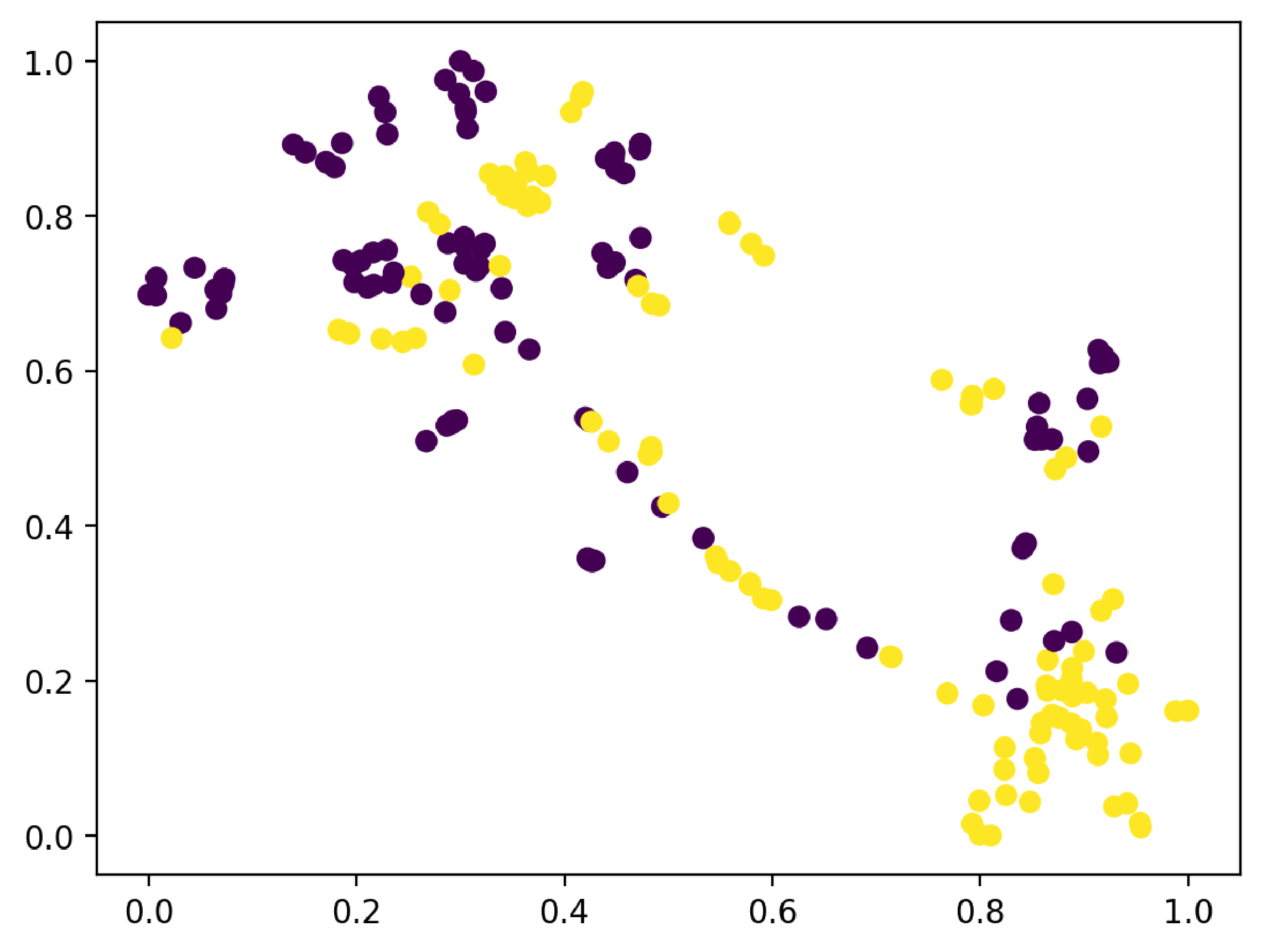}}
			\centerline{(b)}
		\end{minipage}
		\begin{minipage}{0.16\columnwidth}
			\centering
			\centerline{
				\label{fig:5-c}
				\includegraphics[width=\columnwidth]{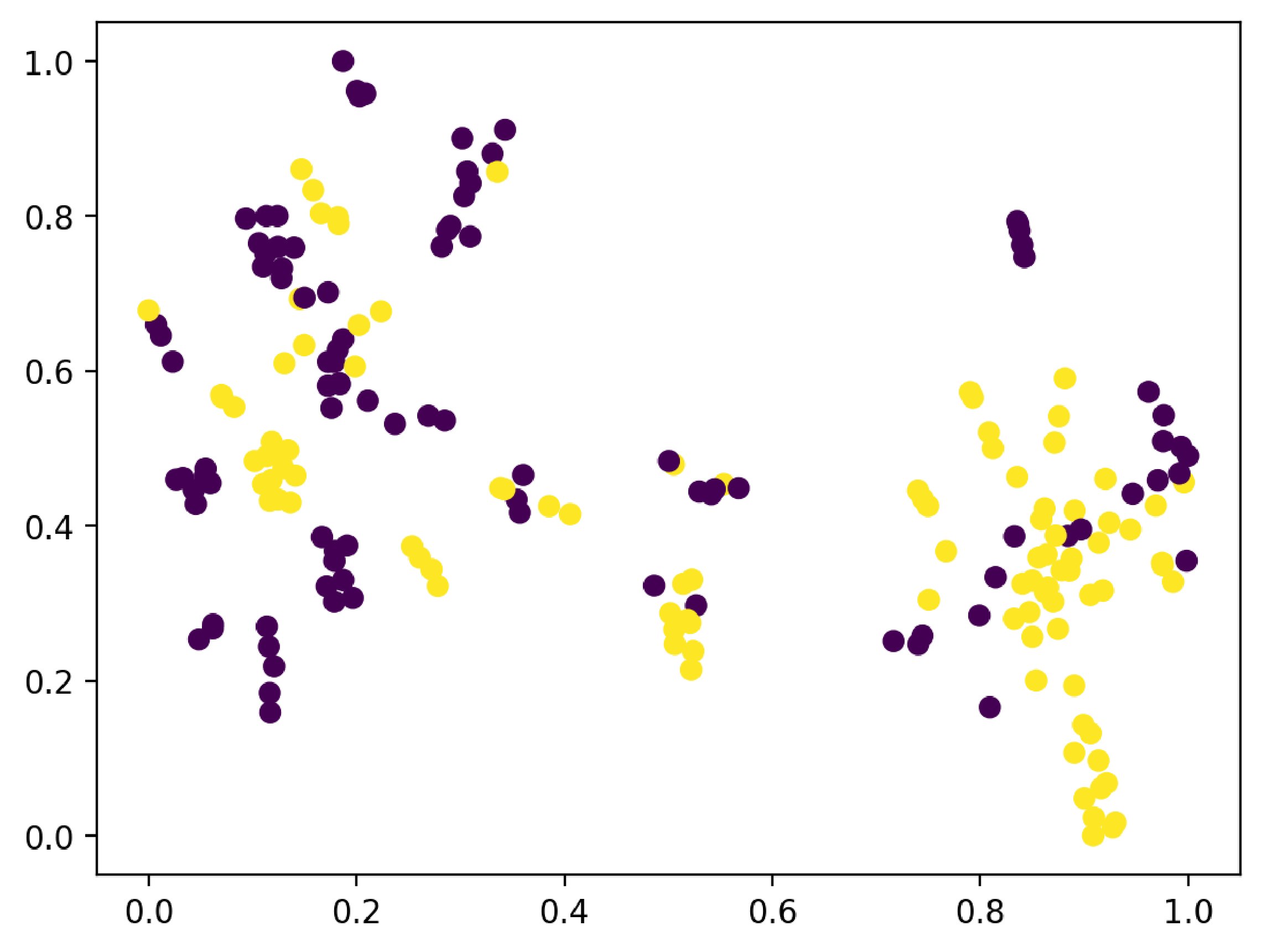}}
			\centerline{(c)}
		\end{minipage}
		\begin{minipage}{0.16\columnwidth}
			\centerline{
				\label{fig:5-d}
				\includegraphics[width=\columnwidth]{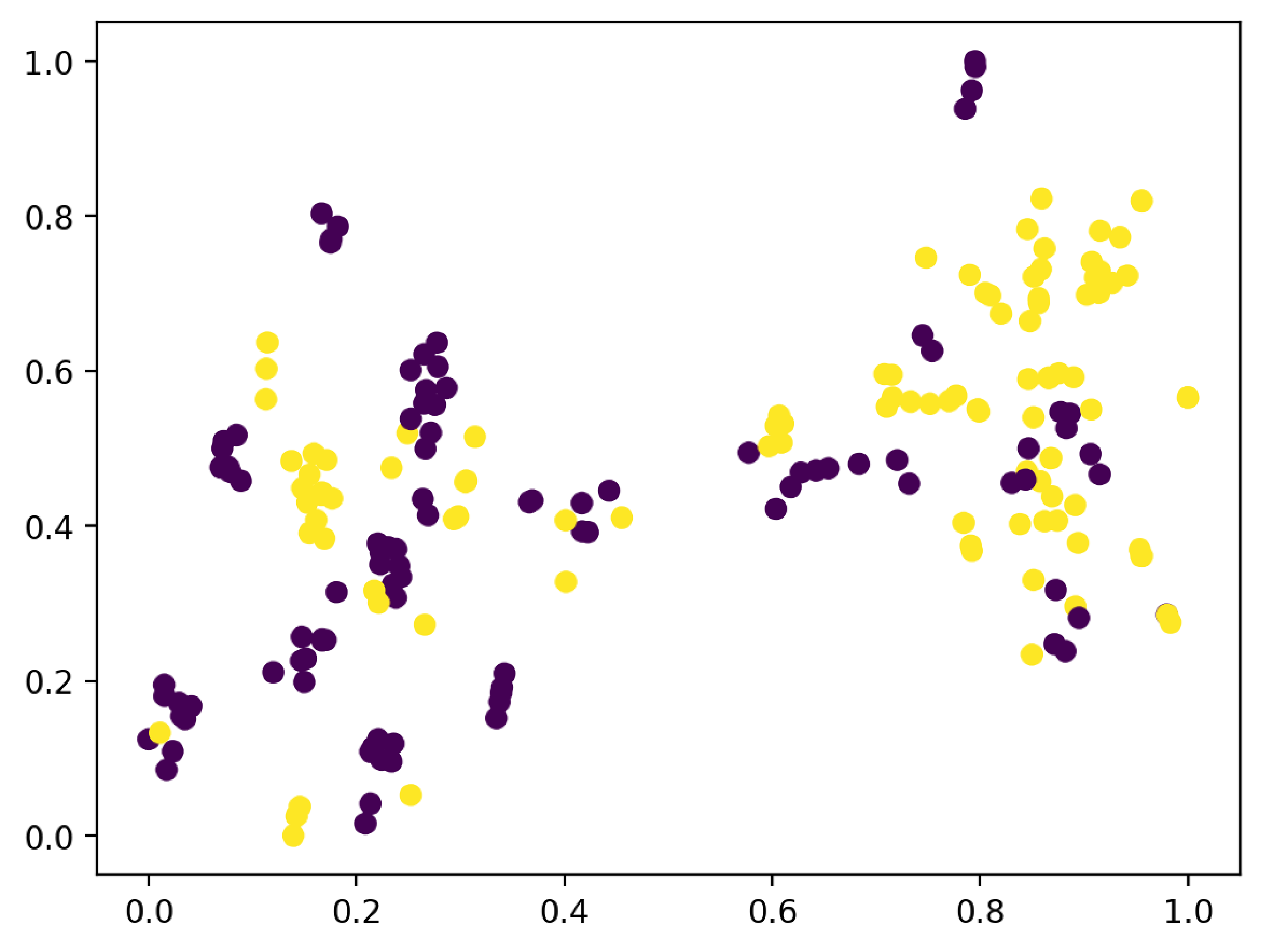}}
			\centerline{(d)}
		\end{minipage}
		\begin{minipage}{0.16\columnwidth}
			\centerline{
				\label{fig:5-e}
				\includegraphics[width=\columnwidth]{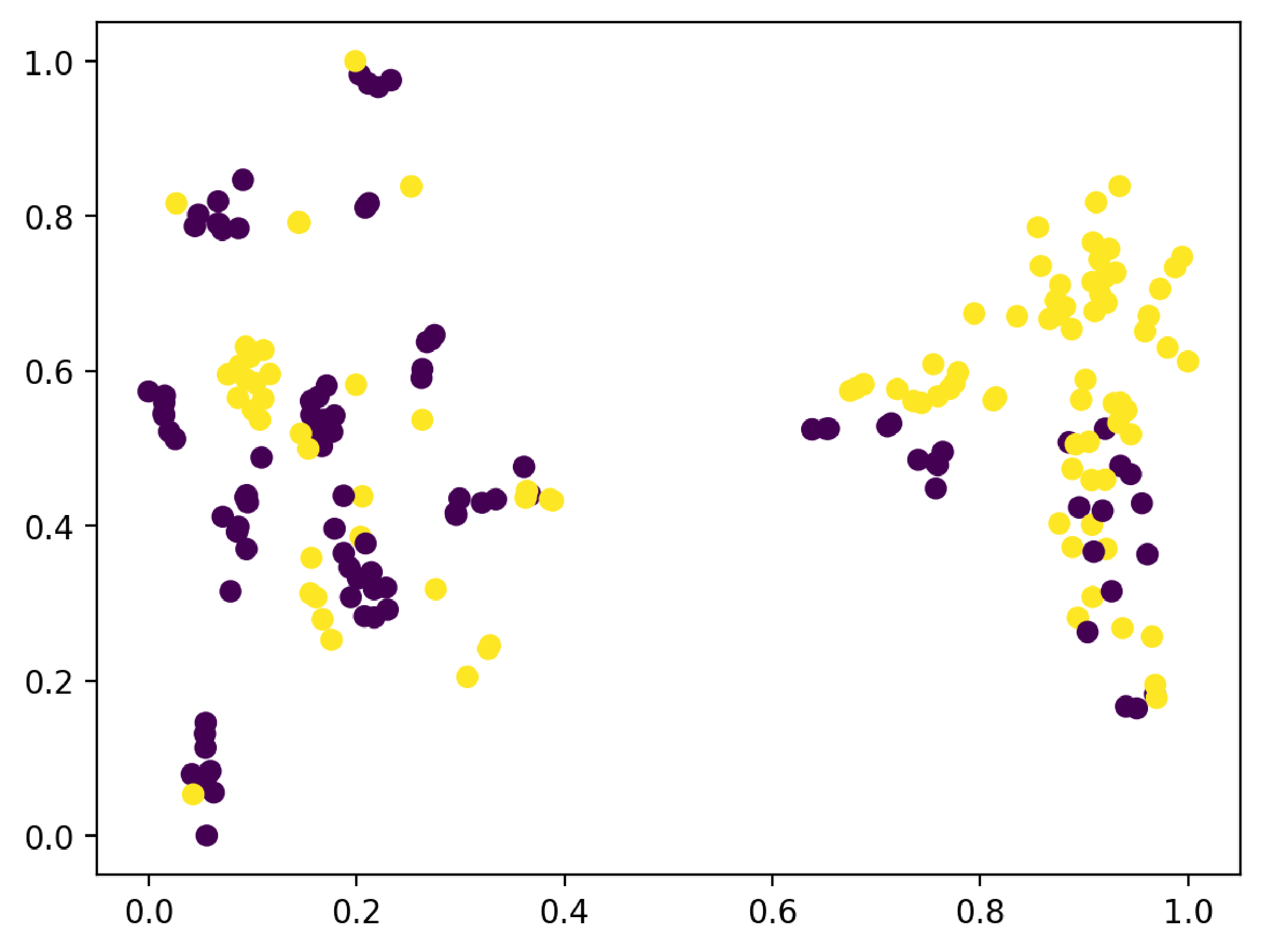}}
			\centerline{(e)}
		\end{minipage}
		\begin{minipage}{0.16\columnwidth}
			\centerline{
				\label{fig:5-f}
				\includegraphics[width=\columnwidth]{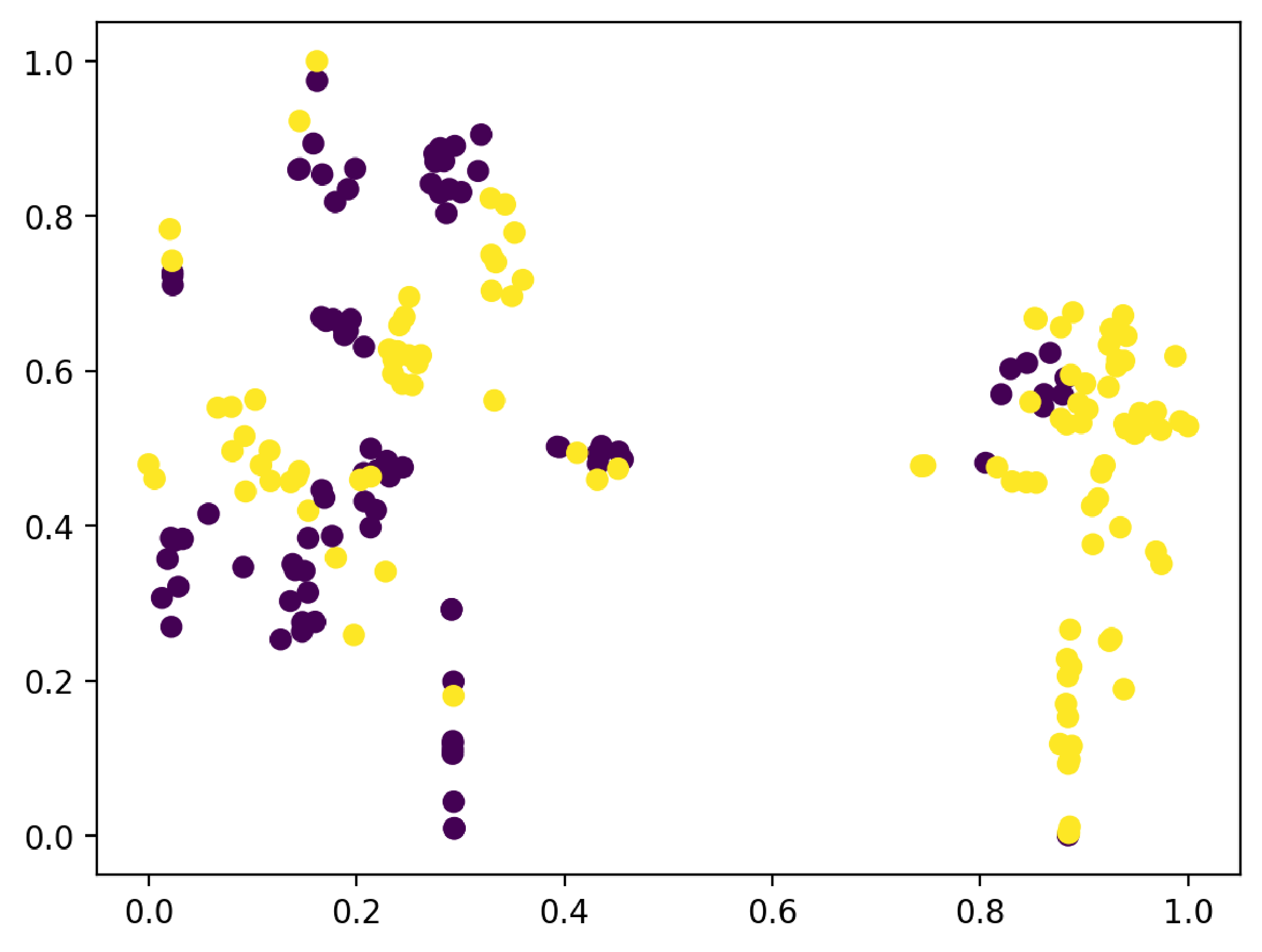}}
			\centerline{(f)}
		\end{minipage}
		\quad   
		\begin{minipage}{0.16\columnwidth}
			\centerline{
				\label{fig:5-g}
				\includegraphics[width=\columnwidth]{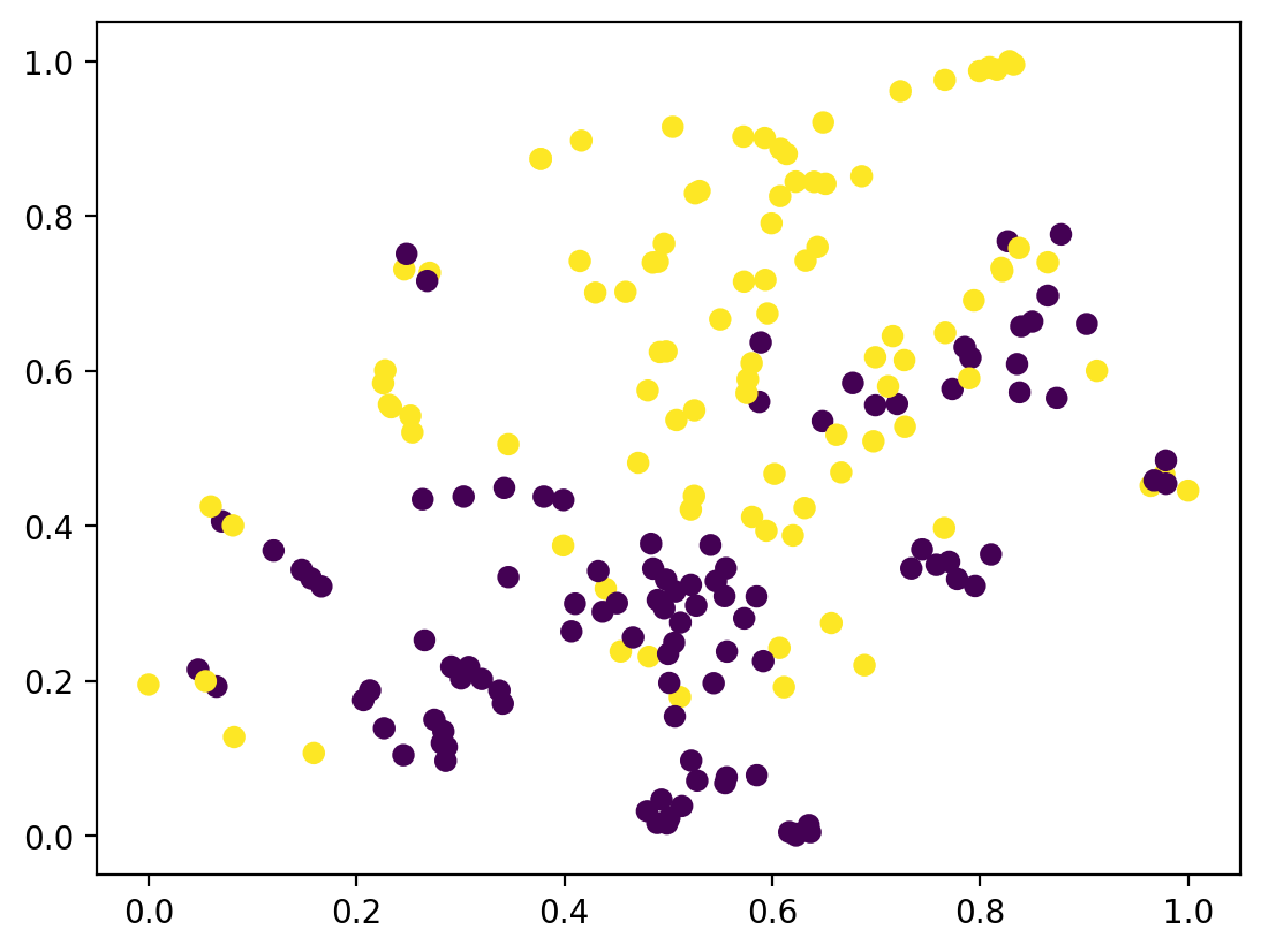}}
			\centerline{(g)}
		\end{minipage}
		\begin{minipage}{0.16\columnwidth}
			\centerline{
				\label{fig:5-h}
				\includegraphics[width=\columnwidth]{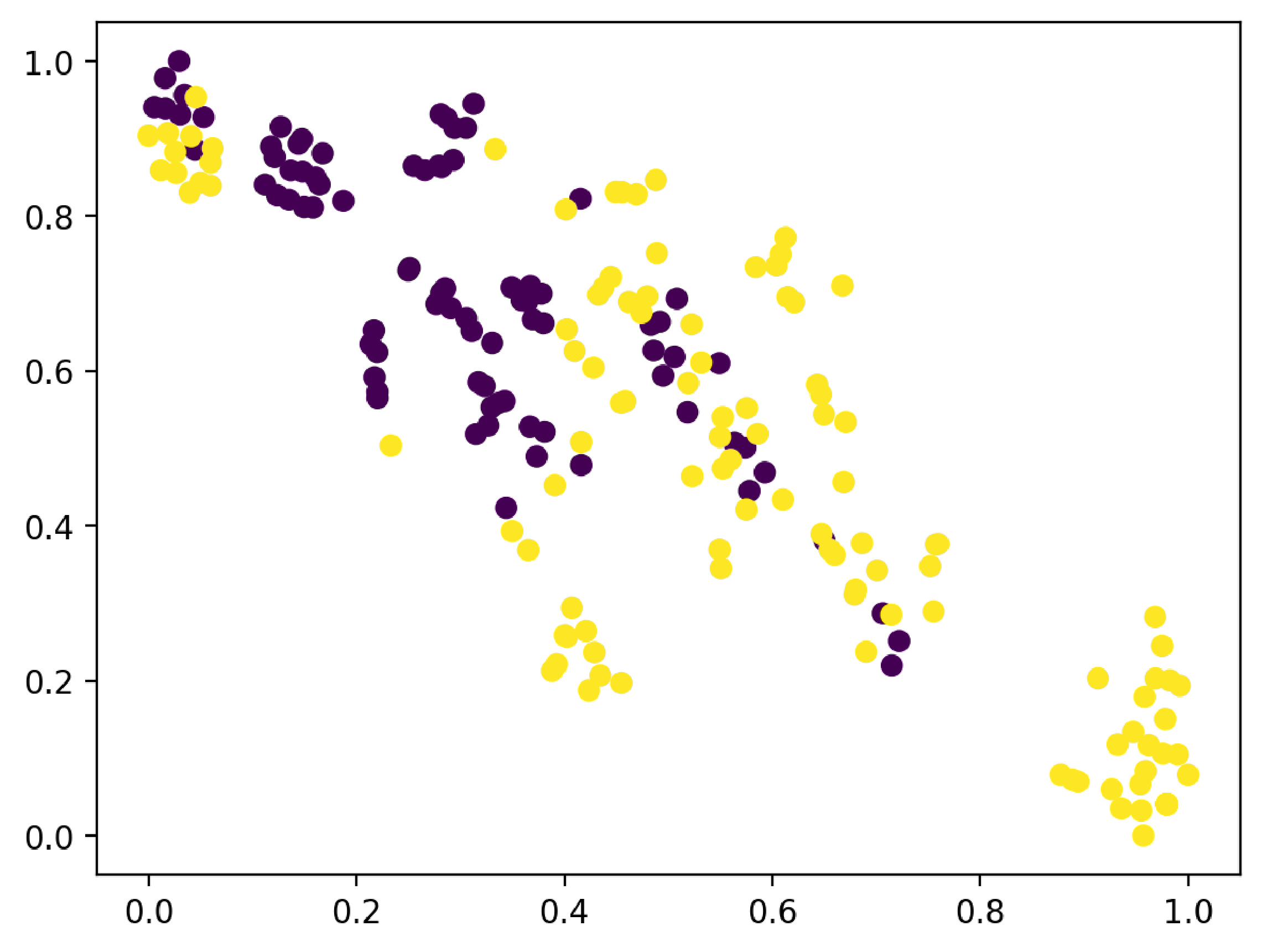}}
			\centerline{(h)}
		\end{minipage}
		\begin{minipage}{0.16\columnwidth}
			\centering
			\centerline{
				\label{fig:5-i}
				\includegraphics[width=\columnwidth]{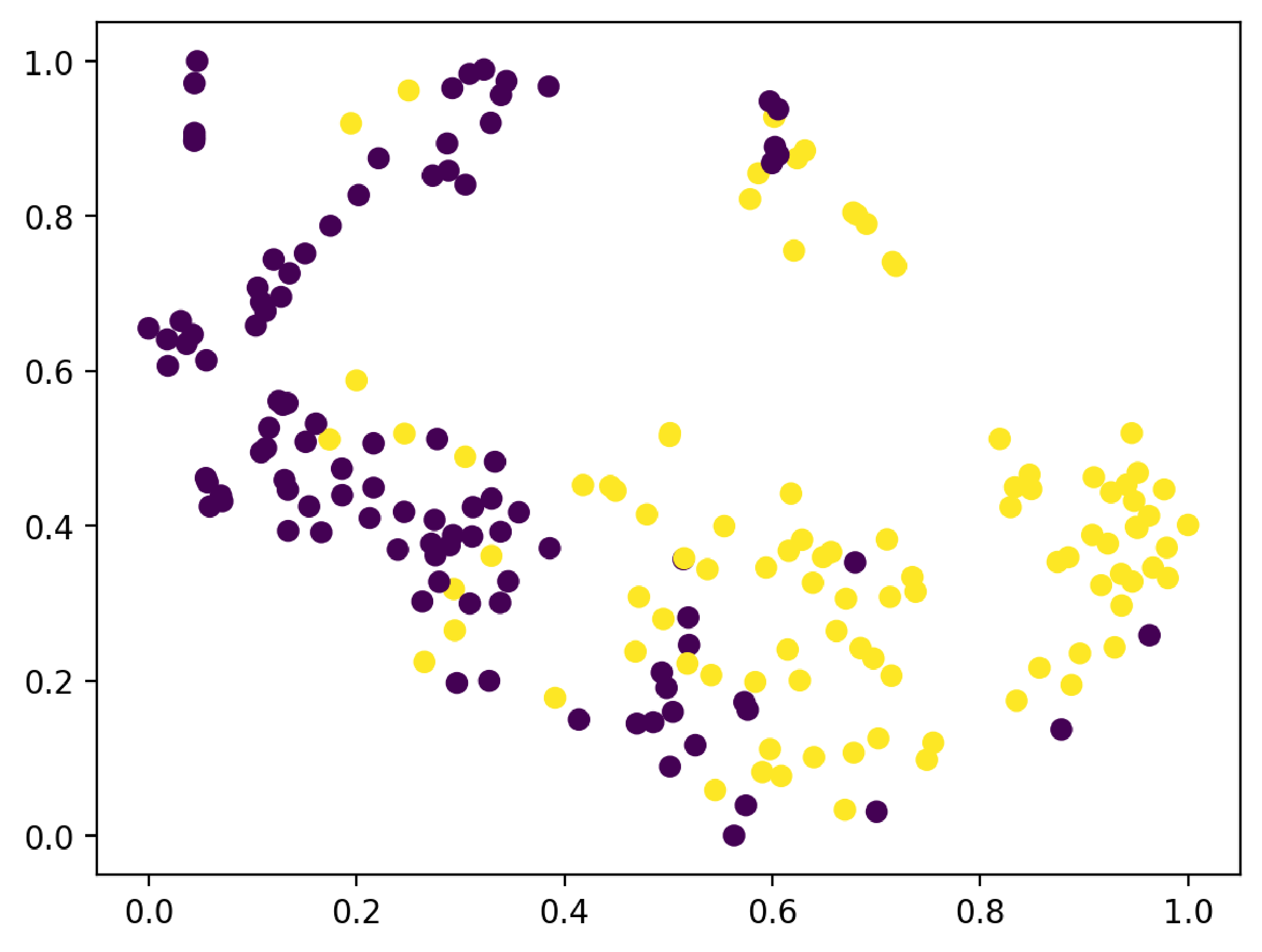}}
			\centerline{(i)}
		\end{minipage}
		\begin{minipage}{0.16\columnwidth}
			\centerline{
				\label{fig:5-j}
				\includegraphics[width=\columnwidth]{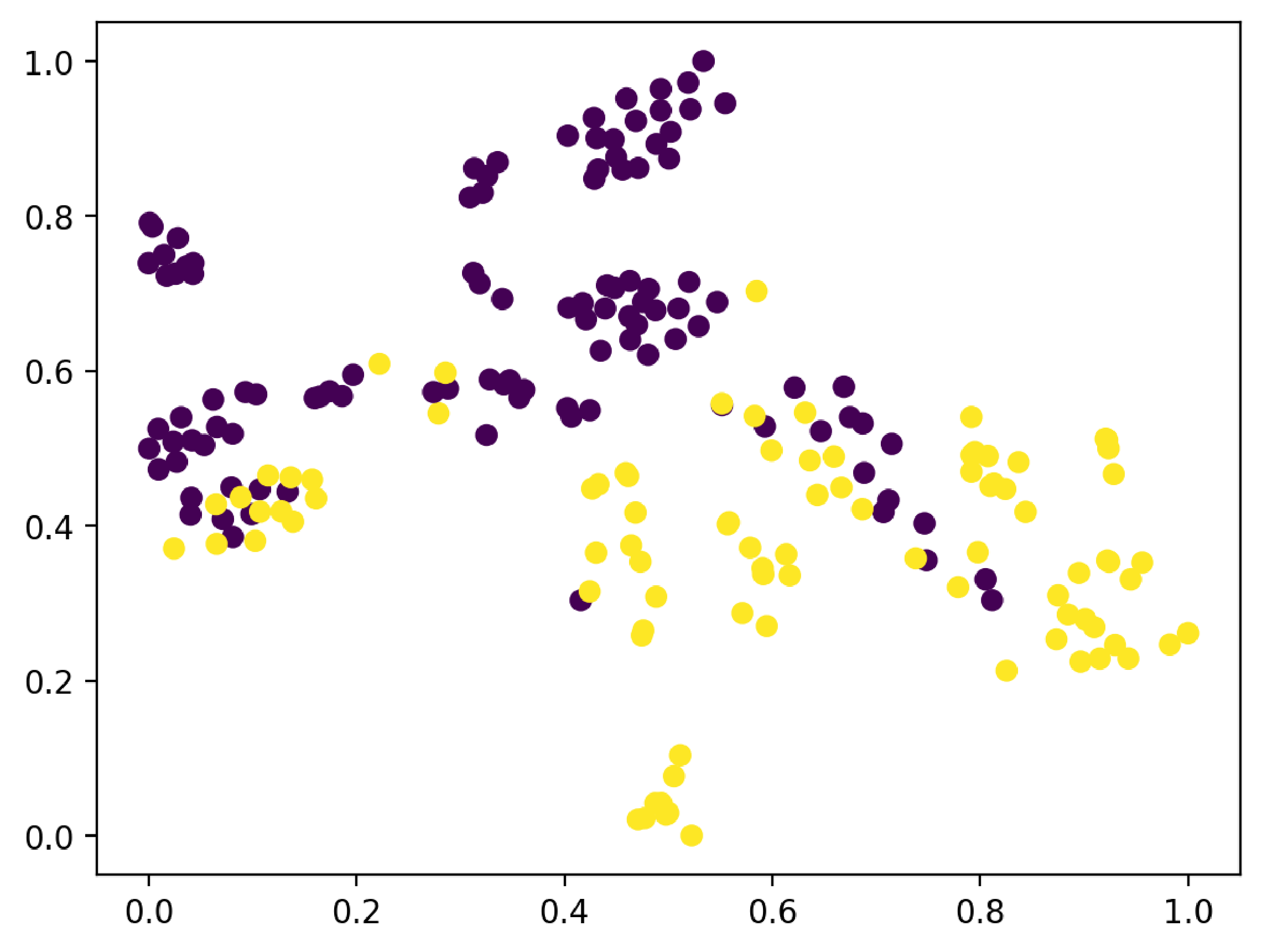}}
			\centerline{(j)}
		\end{minipage}
		\begin{minipage}{0.16\columnwidth}
			\centerline{
				\label{fig:5-k}
				\includegraphics[width=\columnwidth]{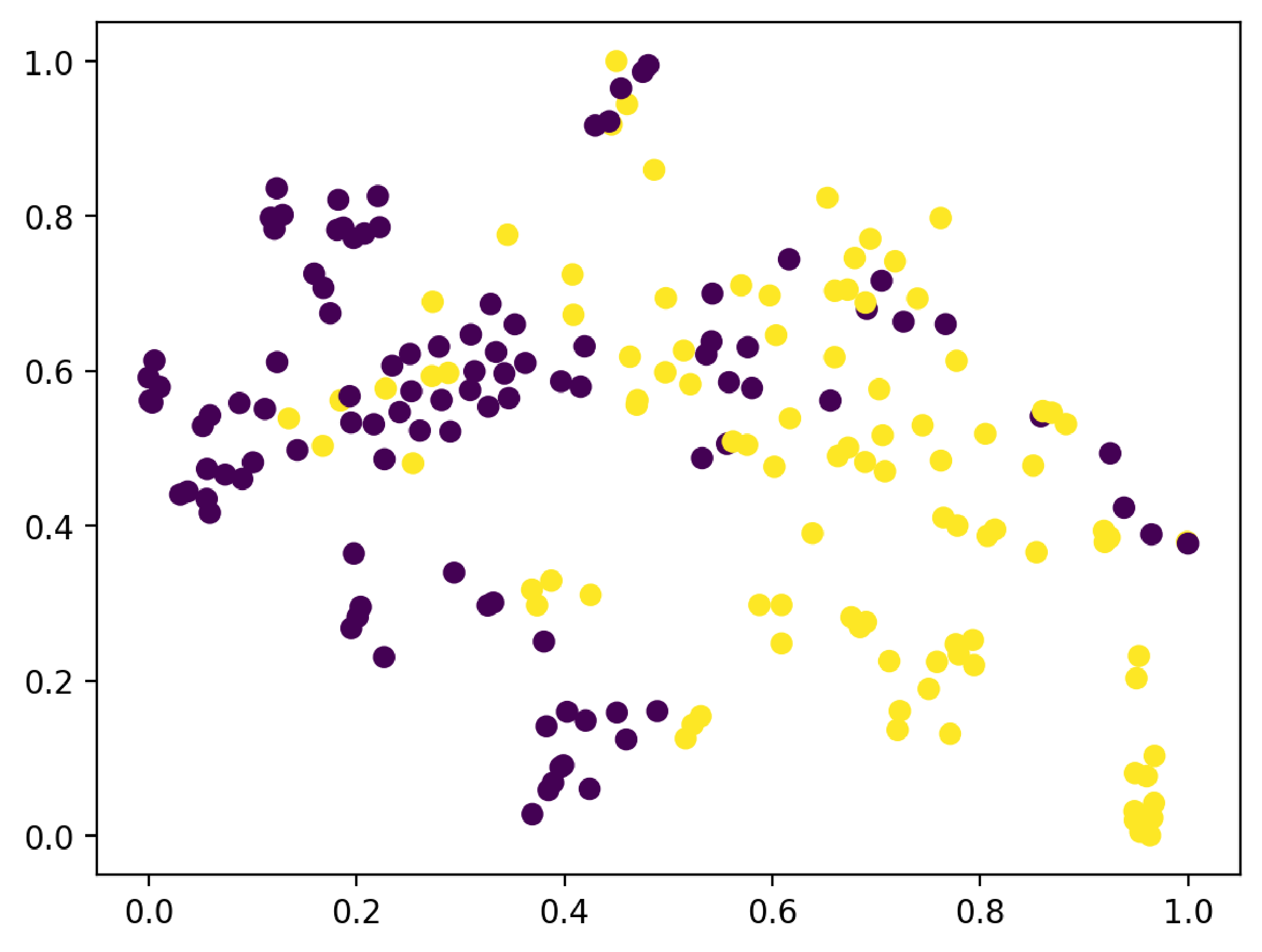}}
			\centerline{(k)}
		\end{minipage}
		\begin{minipage}{0.16\columnwidth}
			\centerline{
				\label{fig:5-l}
				\includegraphics[width=\columnwidth]{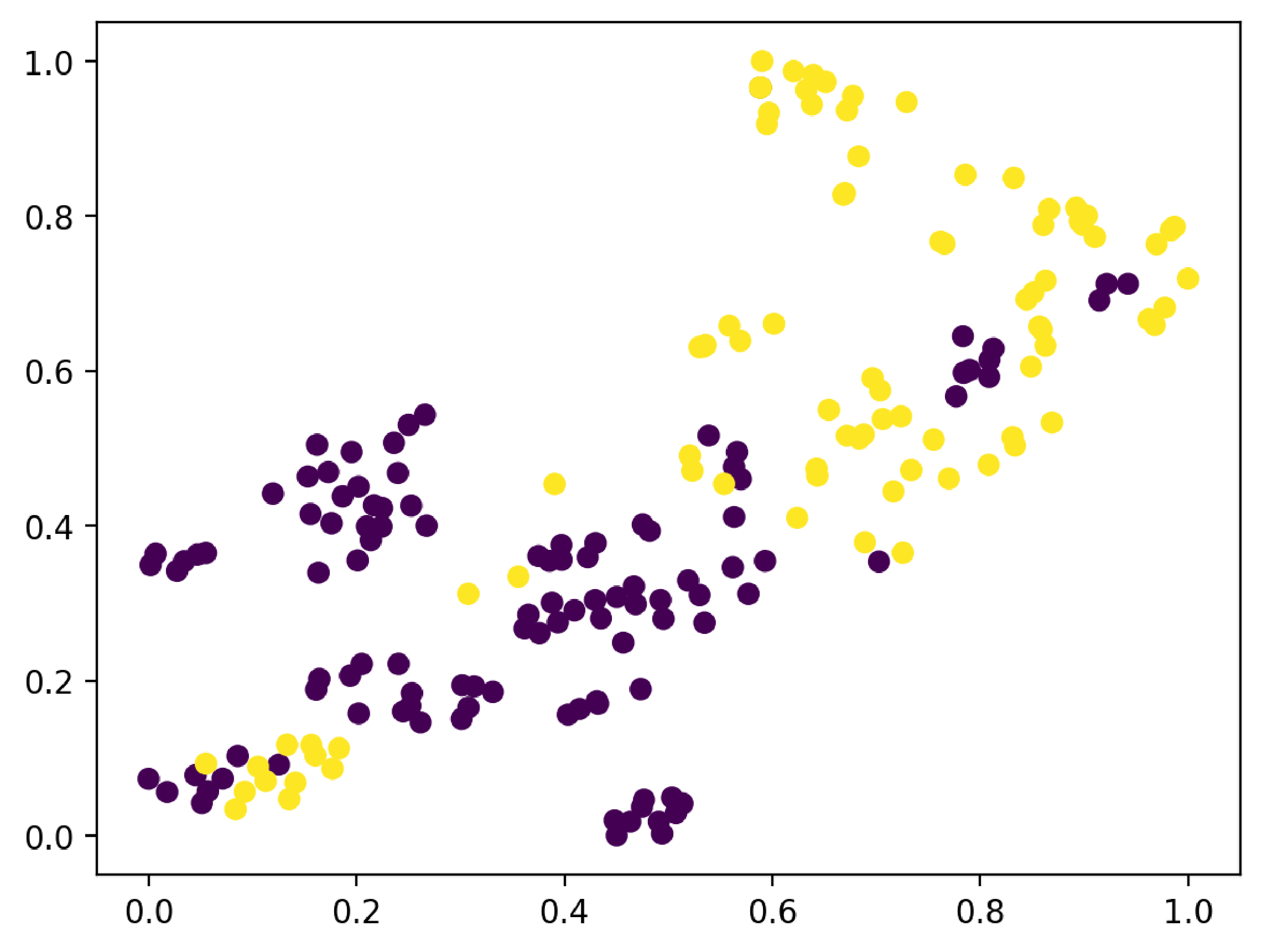}}
			\centerline{(l)}
		\end{minipage}
		\quad   
		\caption{T-SNE visualization of feature representations on BP4D and DISFA datasets. Colors indicates whether AU1 exists. Top row: (a) Representations optimized by WCE on BP4D dataset, (b)-(f) Visualizations of ablation study on BP4D dataset, respectively represent model A-E. Bottom row: (g) Representations optimized by AUNCE on DISFA dataset, (h)-(l) Visualizations of ablation study on DISFA dataset, respectively represent model A-E.} 
		\label{fig:5}
	\end{center}
\end{figure}

To qualitatively assess the efficacy of our discriminative contrastive learning paradigm, we visualize the learned feature representations optimized by AUNCE and WCE using T-SNE dimensionality reduction, as shown in Fig.\ref{fig:5}. A random sample of 200 images from each dataset is used for illustration. Comparing Fig.\ref{fig:5}(a), Fig.\ref{fig:5}(f) with Fig.\ref{fig:5}(g), Fig.\ref{fig:5}(l) demonstrates that features learned with AUNCE show greater separability than those optimized by WCE. Additionally, Fig.\ref{fig:5}(b)-(f) and Fig.~\ref{fig:5}(g)-(l) highlight the contribution of each component in our ablation study.

The ablation studies confirm that the proposed discriminative contrastive learning paradigm enhances the encoder's feature representation capability, addresses class imbalance for each AU type, and improves robustness to noisy and false labels, ultimately boosting generalization performance in AU detection. Each component contributes to performance gains, with NS proving most effective. Furthermore, as shown in Table~\ref{tab:ablation}, the addition of $w_{i}$, PS, and NS consistently improves or maintains performance across nearly all AUs.

\subsubsection{$p_{1}, p_{2}, p_{3}, p_{4}$, the probabilities of various positive sample pairs} 
In our paper, $p_{1}$ represents the probabilities of images with the highest similarity to original images, $p_{2}$ represents the probabilities of self-supervised images enhanced by simple augmentations, $p_{3}$ represents the probabilities of a mixture of all positive samples, and $p_{4}$ represents the probabilities of images with the lowest similarity to original images. Given that $p_{1}$, $p_{2}$, $p_{3}$, and $p_{4}$ address the problem of AU labels tainted by noise and errors, we systematically vary these probabilities from 0 to 1 to investigate their impact on our framework. As $p_{1}$ and $p_{2}$ are deemed reliable positive samples, we set their values to be the same during the variation.

The experimental results are presented in Table~\ref{tab:Hyperparameters Analysis}. When $p_{1}$=0.15, $p_{2}$=0.15, $p_{3}$=0.7, and $p_{4}$=0, the experimental result is optimal. It is observed that images with the lowest similarity to original images contribute minimally to the training process, as these images essentially represent noise and errors. The mixture of all positive samples plays a pivotal role, as it encourages the feature representation of each instance to be closer to the class centroid, facilitating the capture of essential characteristics of the positive class. The indispensability of the remaining two positive samples lies in their ability to mitigate the impact of noisy and false samples, thereby reinforcing the robustness of the training process.

However, hyperparameter tuning is a limitation of our paper. Although the proposed method yields promising results, we observed that the performance is highly sensitive to the values of $p_{1}$, $p_{2}$, and $p_{3}$, with their optimal settings varying across different datasets. Moreover, the parameter tuning process can only be guided by experience gained from extensive experimentation, which results in a tuning scheme that lacks flexibility. 

\subsection{Hyperparameters Analysis} \label{Hyperparameters Analysis}
\begin{table}[htbp]
	\begin{center}
		\scriptsize
		\caption{Quantitative results the probabilities of various positive sample pairs $p_{1}, p_{2}, p_{3}, p_{4}$ on BP4D dataset.}
		\label{tab:Hyperparameters Analysis}
		\renewcommand{\arraystretch}{0.8}
		\tabcolsep=0.5cm
		\begin{tabular}{c c c c c c}
			\toprule[2pt]
			\multicolumn{1}{c}{}&$\textbf{$p_{1}$}$&$\textbf{$p_{2}$}$&$\textbf{$p_{3}$}$&$\textbf{$p_{4}$}$&$\textbf{F1-Score(\%)}$\\ 
			\midrule[1pt]
			\multicolumn{1}{l}{\multirow{8}{*}{$\textbf{Probabilities}$}}&1&0&0&0&60.1\\
			\multicolumn{1}{c}{}&0&1&0&0&48.4\\
			\multicolumn{1}{c}{}&0&0&1&0&63.2\\
			\multicolumn{1}{c}{}&0&0&0&1&17.6\\
			\multicolumn{1}{c}{}&0.1&0.1&0.8&0&63.8\\
			\multicolumn{1}{c}{}&0.15&0.15&0.7&0&$\textbf{66.1}$\\
			\multicolumn{1}{c}{}&0.2&0.2&0.6&0&65.7\\
			\multicolumn{1}{c}{}&0.25&0.25&0.5&0&65.2\\
			\multicolumn{1}{c}{}&0.3&0.3&0.4&0&64.3\\
			\multicolumn{1}{c}{}&0.4&0.4&0.2&0&58.8\\
			\bottomrule[2pt]
		\end{tabular}
	\end{center}
\end{table}

\subsubsection{Backpropagation rate controller $\beta$} \label{beta}
As is well understood, the hyperparameter $\beta$ plays a crucial role in governing the relative importance of minority and majority class samples during the parameter update process. It ensures that the model allocates more attention to minority class samples when necessary and balances the gradient magnitudes accordingly. To explore its effects, we systematically vary the hyperparameter $\beta$ for the minority class from 0.8 to 1.8, with $\beta$ for the majority class fixed at 0.4. Additionally, we vary $\beta$ for the majority class from 0.2 to 1.2, keeping $\beta$ for the minority class fixed at 1.2.

\begin{figure}[htbp] 
	\begin{center}
		\setlength{\abovecaptionskip}{0.05cm}
		\begin{minipage}{0.49\columnwidth}
			\centering
			\centerline{\includegraphics[width=\columnwidth]{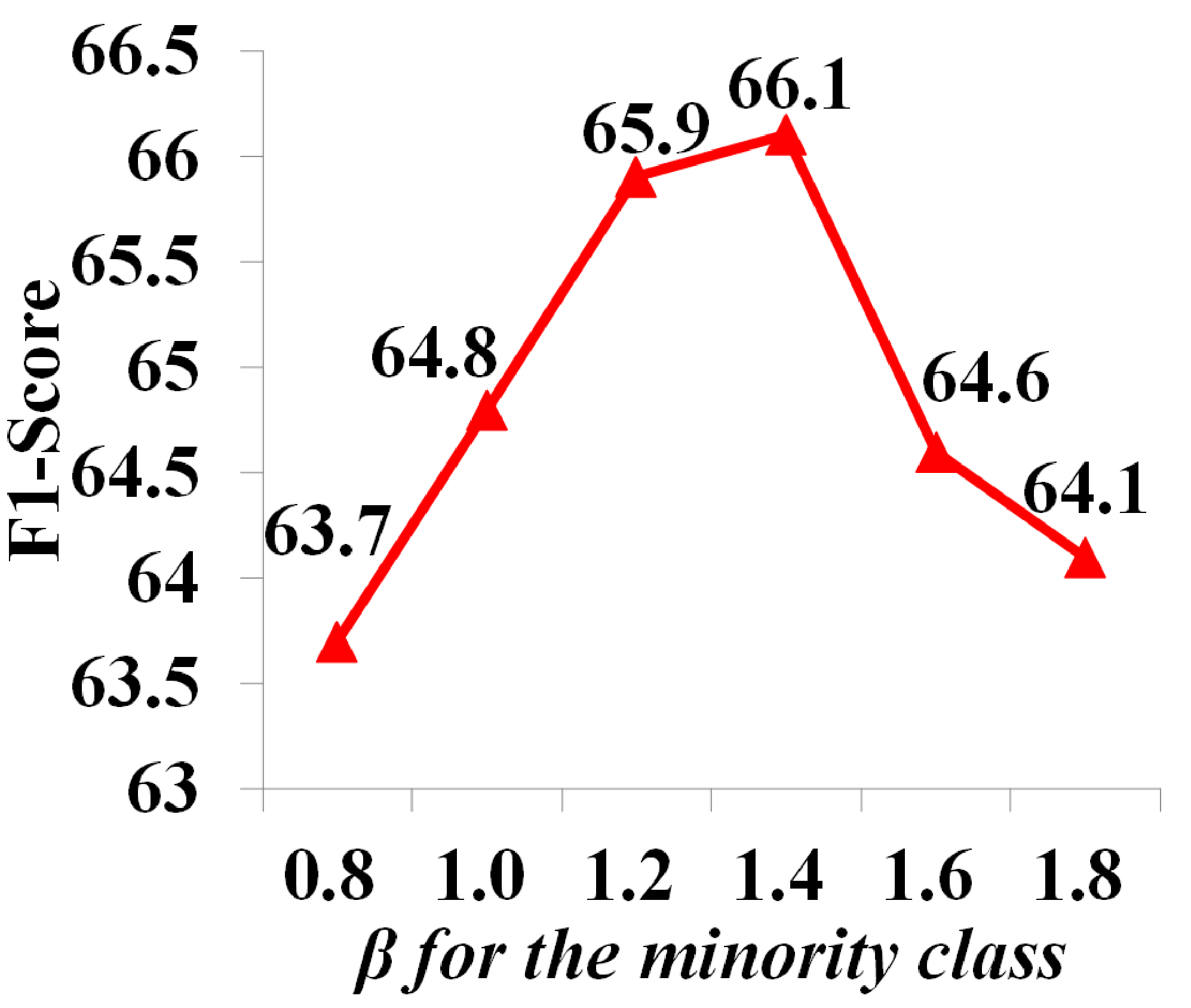}}
			\centerline{(a)}
		\end{minipage}
		\hfill
		\begin{minipage}{0.49\columnwidth}
			\centering
			\centerline{\includegraphics[width=\columnwidth]{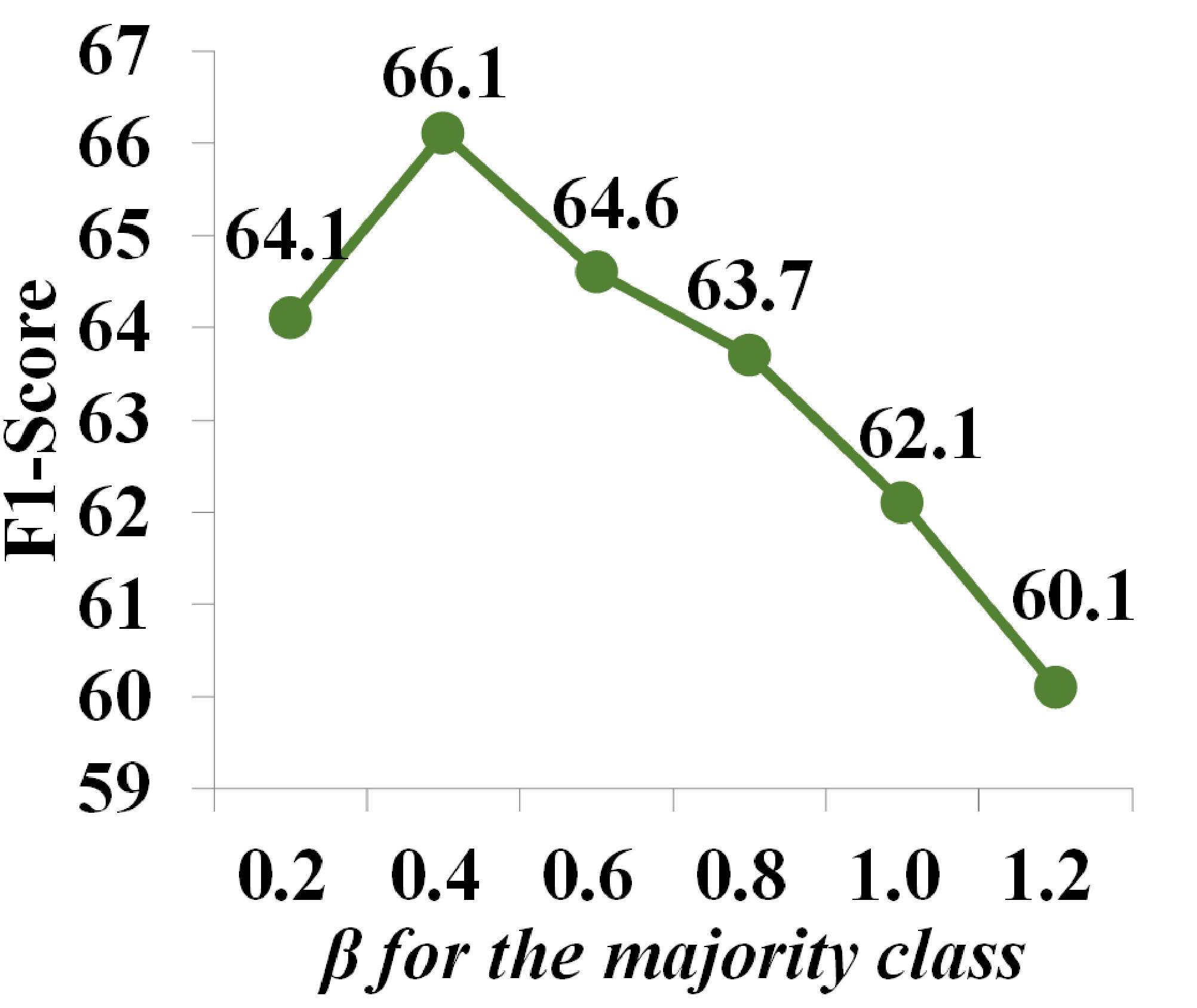}}
			\centerline{(b)}
		\end{minipage}
		\caption{Quantitative results of $\beta$ of different positive and negative samples for each AU type on BP4D dataset. From left to right: (a) $\beta$ for positive samples (the minority class), and (b) $\beta$ for negative samples (the majority class).}
		\label{fig:6}
	\end{center}
\end{figure}

The experimental results are depicted in Fig.~\ref{fig:6}. The optimal experimental outcome is observed when $\beta$ for the minority class is set to 1.2, and $\beta$ for the majority class is fixed at 0.4. Hence, when $\beta_{\text{minority}} > \beta_{\text{majority}}$, the impact of the class imbalance issue can be effectively mitigated.

\section{Conclusion}
In this paper, we propose a discriminative contrastive learning framework tailored for AU detection, introducing a novel contrastive loss named AUNCE. Building upon the InfoNCE loss, AUNCE refines the task by shifting focus from learning pixel-level information of the entire face to capturing subtle feature variations during AU activation, thus improving detection efficiency. We also introduce a positive sample sampling strategy that constructs three types of positive sample pairs, selectively augmenting positive samples, especially in the presence of noisy and false AU labels. This approach ensures the model is trained on more reliable examples, improving upon conventional methods that lack adaptive sampling. To address class imbalance in AU detection, we implement a negative sample re-weighting strategy that prioritizes minority AUs, helping the model focus on hard-to-detect samples. This strategy promotes hard example mining by giving more weight to underrepresented classes, enhancing model performance and generalization. Experimental results demonstrate that AUNCE consistently outperforms existing methods, particularly those utilizing weighted cross-entropy loss.

Despite the overall effectiveness of our approach, hyperparameter selection remains a limitation. Model performance is sensitive to certain parameters, with optimal configurations varying across datasets. The current tuning process is largely empirical and lacks generalizability. Future work will focus on developing adaptive hyperparameter optimization strategies to enhance robustness and generalization across diverse tasks. In addition, we plan to incorporate recently proposed loss functions designed to address class imbalance into the AU detection task. These include techniques based on class reweighting, margin adjustment, and adaptive sampling, which have shown promise in improving performance on underrepresented classes and enhancing overall model robustness.

\section*{Acknowledgement}
This research was supported by the Fundamental Research Funds for the Central Universities (No.2682025CX082), the National Natural Science Foundation of China (Nos. 62176221, 61572407), and Sichuan Province Science and Technology Support Program (Nos. 2024NSFTD0036, 2024ZHCG0166). 

\section*{Declaration of generative AI and AI-assisted technologies in the writing process}
During the preparation of this work the author(s) used $chatgpt$ in order to polish our paper. After using this tool/service, the author(s) reviewed and edited the content as needed and take(s) full responsibility for the content of the publication.

\bibliographystyle{elsarticle-num}
\bibliography{egbib}
\end{document}